\documentclass[pdflatex,sn-mathphys-num]{sn-jnl}% Math and Physical Sciences Numbered Reference Style
\usepackage{graphicx}%
\usepackage{multirow}%
\usepackage{amsmath,amssymb,amsfonts}%
\usepackage{amsthm}%
\usepackage{mathrsfs}%
\usepackage[title]{appendix}%
\usepackage{xcolor}%
\usepackage{textcomp}%
\usepackage{manyfoot}%
\usepackage{booktabs}%
\usepackage{algorithm}%
\usepackage{algorithmicx}%
\usepackage{algpseudocode}%
\usepackage{listings}%
\usepackage[thinc]{esdiff}
\usepackage{bbm}
\theoremstyle{thmstyleone}%
\theoremstyle{thmstyletwo}%

\theoremstyle{thmstylethree}%

    \newcommand{\beginsupplement}{%
        \setcounter{table}{0}%
        \renewcommand{\thetable}{S\arabic{table}}%
        \setcounter{figure}{0}%
        \renewcommand{\thefigure}{S\arabic{figure}}%
        \setcounter{equation}{0}%
        \renewcommand{\theequation}{S\arabic{equation}}%
    }
\begin{document}

\title[Article Title]{When Routers, Switches and Interconnects Compute: A Processing-in-Interconnect Paradigm for Scalable Neuromorphic AI}


\author[1]{\fnm{Madhuvanthi} \sur{Srivatsav R}}\email{madhuvanthis@iisc.ac.in}

\author[2]{\fnm{Chiranjib} \sur{Bhattacharyya}}\email{chiru@iisc.ac.in}

\author*[3]{\fnm{Shantanu} \sur{Chakrabartty}}\email{shantanu@wustl.edu}

\author*[1]{\fnm{Chetan} \sur{Singh Thakur}}\email{csthakur@iisc.ac.in}

\affil[1]{\orgdiv{Department of Electronic Systems and Engineering}, \orgname{Indian Institute of Science}, {\city{Bangalore}, \country{India}}}

\affil[2]{\orgdiv{Department of Computer Science and Automation}, \orgname{Indian Institute of Science}, {\city{Bangalore}, \country{India}}}

\affil[3]{\orgdiv{Department of Electrical and Systems Engineering}, \orgname{Washington University}, \orgaddress{ \state{St. Louis}, \country{USA}}}


\abstract{ 
Routing, switching, and the interconnect fabric are essential components in implementing large-scale neuromorphic computing architectures. While this fabric plays only a supporting role in the process of computing, for large AI workloads, this fabric ultimately determines the overall system's performance, such as energy consumption and speed. In this paper, we offer a potential solution to address this bottleneck by addressing two fundamental questions: (a) What computing paradigms are inherent in existing routing, switching, and interconnect systems, and how can they be used to implement a Processing-in-Interconnect ($\pi^2$) computing paradigm? and {(b) how to train $\pi^2$ network on standard AI benchmarks ?} To address the first question, we demonstrate that all operations required for typical AI workloads can be mapped onto delays, causality, time-outs, packet-drop, and broadcast operations, all of which are already implemented in current packet-switching and packet-routing hardware. {We then show that existing buffering and traffic-shaping embedded algorithms can be minimally modified to implement $\pi^2$ neuron models and synaptic operations. To address the second question, we show how a knowledge distillation framework can be used to train and cross-map well-established neural network topologies onto $\pi^2$ architectures without any degradation in the generalization performance. Our analysis show that the effective energy utilization of a $\pi^2$ network is significantly higher than that of other neuromorphic computing platforms, as a result we believe that the $\pi^2$ paradigm offers a more scalable architectural path toward achieving brain-scale AI inference.}
}

\keywords{Routing, Switching, Interconnects, neuromorphic computing, artificial intelligence, Network simulators}



\maketitle

\section{Introduction}\label{sec1}

While advances in artificial intelligence and neuromorphic hardware accelerators are being driven by faster and more energy-efficient computing platforms~\cite{h100,tpu,tpuv6,nvidiacpu,nvidiacpu1,npu,amdfpga,andre} and integrated memory technology~\cite{tsun,hbm_pim}, 
%platforms— such as graphical processing units (GPUs) \cite{h100}, tensor processing units (TPUs) \cite{tpu,tpuv6}, central processing units (CPUs) \cite{nvidiacpu,nvidiacpu1}, neural processing units (NPUs) \cite{npu}, and field-programmable gate arrays (FPGAs) \cite{amdfpga}, 
an often underappreciated trend is the rapid evolution of switching and interconnect technology. Modern Ethernet switches now deliver aggregate data rates exceeding 51.2 Tera-bits-per-second with energy-efficiencies better than 10pJ/bit \cite{marvel_tera}, with projections forecasting performance reaching 4.096 Peta-bits-per-second at 1pJ/bit by 2034 \cite{cisco}. These high-throughput routing fabrics, whether integrated on heterogeneous packaging substrates \cite{truenorth_large_scale} or deployed as standalone switching units \cite{cisco_one_family}, are essential for scaling AI and neuromorphic systems to brain-scale workloads and beyond \cite{large_scale_neuro}. Despite their critical role, interconnects increasingly pose a bottleneck to the overall system performance and energy efficiency. They introduce latency \cite{latency}, jitter \cite{jitter}, and congestion-induced packet loss, resulting in unpredictable behavior and degraded system reliability \cite{photonic}. More significantly, interconnects dominate the energy budget in large-scale AI systems, as data movement consumes substantially more energy than computation and memory read/write itself. For instance, transmitting data across a two-dimensional integrated circuit can require up to 80 times more energy than executing an equivalent computation \cite{dendrocentric}. 

\begin{figure*}[!htbp]
\centering
\includegraphics[width=\textwidth]{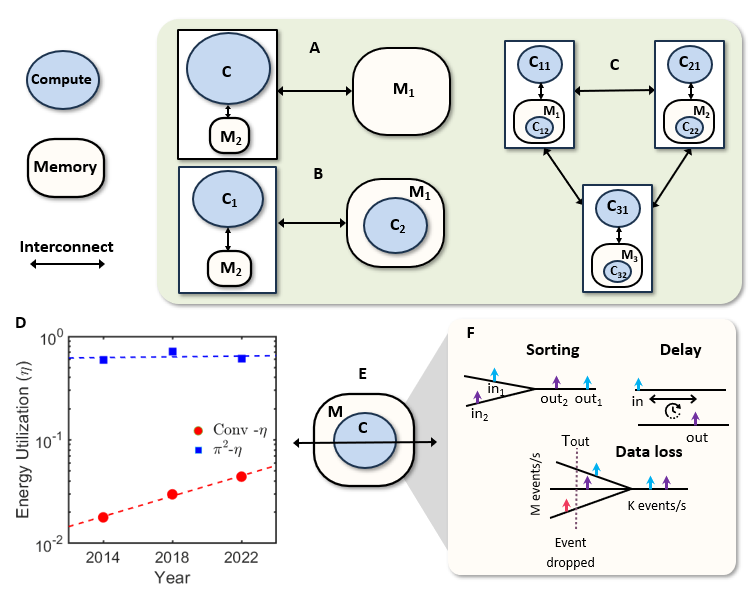}
\caption{A) Standard Von-Neumann architecture where data is moved to and from different memory hierarchies (labeled as $M_1$ and $M_2$) to the compute unit (labeled as C). B) In Compute-in-Memory (CIM) architectures, some compute operations are performed directly within the memory arrays, significantly reducing data movement and associated latency. 
%However, operations that cannot be efficiently mapped to in-memory computation, such as non-linear transformations, often require data to be offloaded to external compute units (CUs), thereby reintroducing data movement and associated overheads. 
C) Distributed neuromorphic architecture with multiple compute and memory cores communicating over the interconnect fabric (routers and switches). D) Projected energy utilization factor $\eta$ for conventional (Conv) and $\pi^2$ architectures with advancement in CMOS technology. 
%across technology nodes - Each data point for a given year, for both architectures, corresponds to a representative CMOS technology node—28nm in 2014, 16nm in 2018, and 7nm in 2022. Conventional architectures exhibit a gradual increase in utilization with node scaling, but remain constrained by interconnect energy bottlenecks and exhibit $\eta$ close to 0. In contrast, $\pi^2$ maintains consistently high effective energy utilization by performing computation within the interconnect. 
E) Processing-in-interconnect ($\pi^2$) paradigm where interconnects serve as memory, compute, and communication units.  F)  $\pi^2$ compute primitives using fundamental interconnect operations: sorting of events (equivalent to ADD), delay of events (equivalent to MULTIPLY and memory) and event time-outs and drops (equivalent to non-linear activation). 
%The inherent properties of the interconnects enable it to i) sort events in the time domain, ii) delay events by virtue of their inherent latency in communication, iii) drop the events based on some thresholding function or due to congestion, thus imbibing non-linearity in computation. 
}
\label{fig1}
\end{figure*}

Several computational paradigms have been proposed in the literature to address interconnect bottlenecks, as illustrated in Fig.~\ref{fig1}(A)-(C). Most of these methods physically co-locate compute and memory units, aiming to alleviate bandwidth constraints and to improve energy efficiency.  For example, Fig.\ref{fig1}(B) illustrates a compute-in-memory (CIM) architecture, where matrix-vector operations are performed directly within the memory arrays, significantly reducing data movement \cite{jason,tsun}. While the analog and digital variants of the CIM paradigm can be implemented at wafer-scale~\cite{CEREBRAS}, scaling limitations and the heterogeneous nature of AI workloads require re-partitioning of the compute, CIM, and memory functions into distributed units. The result is a neuromorphic architecture, which is shown in Fig.~\ref{fig1}(C), which comprises of distributed cores with tightly integrated memory and compute units interconnected via a communication fabric \cite{brainscales}. Even in this distributed architecture, the communication fabric acts as the dominant performance and energy bottleneck \cite{large_scale_neuro}.
%In all these architectures, system performance is constrained by the interconnect bottleneck.
For instance, simulating one billion neurons on a large-scale system using TrueNorth processors is projected to consume approximately 4kW of power, with only 300W attributed to computation and the remainder consumed by high-performance networking switches and power supplies \cite{truenorth3}. At brain scale, TrueNorth systems require several hundred kilowatts solely for communication and configuration infrastructure \cite{truenorth_large_scale}. Similarly, simulating one billion neurons—approximately 1\% of the human brain—on a SpiNNaker platform using 50,000 chipsets (each consuming between 0.5 and 0.76W) results in a total system power consumption exceeding 25kW. In SpiNNaker, the power dissipation due to the synaptic operations scales quadratically with the number of neurons and correlates well with inter-core communication activity. Whereas, the power dissipated at the neuron scales only linearly \cite{spinnaker}. To quantify this bottleneck due to communication infrastructure, we define an effective energy utilization metric, $\eta$, as the ratio of energy used for actual computation to the total system energy consumed (including both compute and communication). This metric is shown in Fig.~\ref{fig1}(F) for the architectures shown in Fig.~\ref{fig1}(A)-(C). Details of calculating $\eta$ based on published data can be found in the Methods section~\ref{secM1}. For conventional neuromorphic architectures, even though $\eta$ is projected to increase (due to improved efficiency of the communication links), as shown in Fig.~\ref{fig1}(D), $\eta$ will still be much lower than 1.   
%These limitations underscore the need for architectural paradigms that minimize or repurpose interconnect overheads to mitgate their bottlenecks for networks at scale. 
%These observations underscore that interconnect-related energy consumption will significantly constrain the energy-efficiency of a neuromorphic architecture at scale, especially, if the goal is to exceed, the throughput and the energy-efficiency of the human brain, which can execute an equivalent $10^{16}$ synaptic operations per second at just 20W~\cite{spinnaker2}.  

% In this work, we address the scalability and energy utilization bottleneck by introducing a novel compute paradigm where interconnects function as both memory and processing elements (PEs), as shown in Fig. \ref{fig1}d,e. For processors with passive interconnects, increasing the interconnect bandwidth helps in achieving peak performance faster. It does not in any way contribute to the maximum throughput achieved.  Since transmission energy is traditionally treated as overhead with no computational utility, existing architectures exhibit poor energy proportionality and limited scalability as depicted in Fig. \ref{fig1}G. Addressing this imbalance requires a radical rethinking of the role of interconnects—not just as communication links, but as active computational substrates.

In this paper, we propose a processing-in-interconnect ($\pi^2$) compute architecture, where rather than considering interconnects as an overhead, the interconnect primitives are repurposed and utilized for computing. In this regard, the $\pi^2$ paradigm is neuromorphic in its essence, since biological brains are also hypothesized to exploit axonal and dendritic delays for computing~\cite{dendrocentric,axonal}. However, in $\pi^2$ architecture, our goal is to leverage the interconnect primitives that already exist or are efficiently implemented in current routing, switching, and interconnect hardware. These interconnect primitives are summarized in Fig.~\ref{fig1}F and their equivalence to neural network operations is described in the Methods section ~\ref{secM2}. For instance, the delay operation is shown to be equivalent to a MULTIPLY operation and a memory operation, the sorting operation is shown to be equivalent to an ADD operation, and event time-out/drop-out is shown to be equivalent to the non-linear activation. Since we will be considering only packet switching technologies for implementing the interconnect fabric, events and data packets will be used interchangeably. In this regard, $\pi^2$ architecture would also be able to leverage the event routing infrastructure in existing address-event-representation (AER) neuromorphic infrastructure~\cite{temp,gert,gert2}.
%In traditional neural networks, computations are structured around three primary operations: weighted multiplication, additive integration, and non-linear activation.  In $\pi^2$ network the equivalent operations will be achieved by delaying, time-outs, buffering and packet drops.
% \begin{itemize}
%     \item Delaying introduces temporal shifts that modulate the timing of events
%     \item Time-based sorting reorders events based on relative arrival times, enabling structured integration of inputs
%     \item Event dropping introduces selectivity and thresholding behavior, serving as an additional non-linear mechanism
% \end{itemize} 
Since the computing and the memory functions are integrated with the interconnect functions as shown in Fig.~\ref{fig1}E, the energy-utilization metric $\eta$ for the $\pi^2$ architecture is not constrained by the interconnect bottlenecks. Increasing the interconnect bandwidth effectively scales the computational bandwidth, and since interconnects also serve as memory elements, they alleviate the memory-bandwidth bottlenecks in conventional architectures \cite{cim1,cim2} (Supplementary section ~\ref{secA1} and Fig. \ref{figa1}). As a result, the effective energy-utilization metric $\eta$ for the $\pi^2$ architecture is close to 1, as shown in Fig.~\ref{fig1}F. The key contribution of this work is to show how the $\pi^2$ architecture can be implemented and scaled using packet-switching networks and using well-established network protocols and routing primitives {with minimal modifications}, and how the $\pi^2$ architecture can be trained to execute typical AI tasks. 

%In the following sections, we present the computational dynamics of the $\pi^2$ neuron model and synapse and detail its implementation within the framework of Ethernet-based communication protocols.

%\subsection{ $\pi^2$-CBS: Modeling credit-based traffic shaping algorithm as a neuron Model}

\subsection{Credit-based traffic shaping acts as $\pi^2$ Neuron}
\begin{figure*}[!htbp]
\centering
\includegraphics[width=1\textwidth]{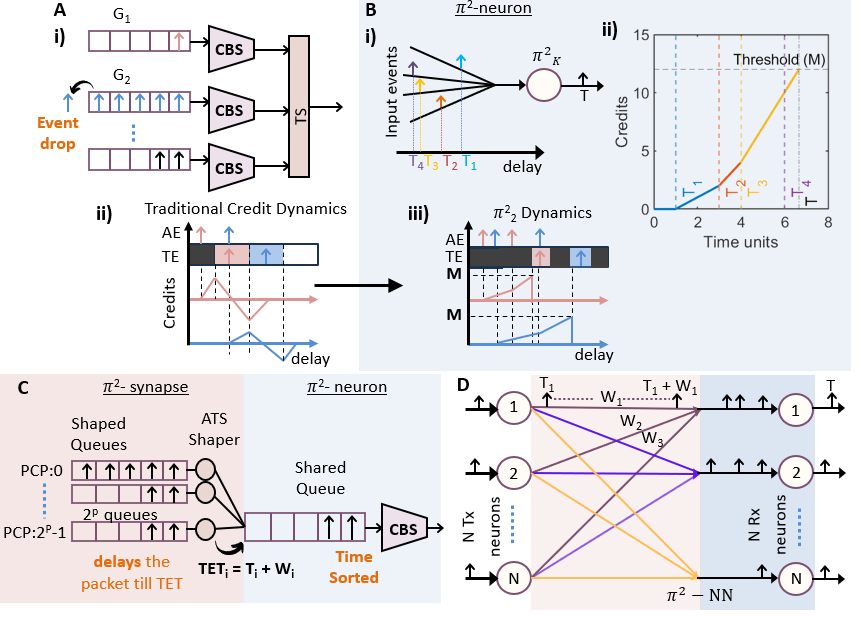}
\caption{ A) Illustration of traditional Credit-Based Shaper (CBS) operation where each traffic class (e.g., $G_1$, $G_2$) is associated with an individual queue and a CBS module. Events are dropped if the queue overflows. Transmission occurs only when the associated credit is non-negative and the channel is free. In ii), Credit dynamics for two traffic classes: $G_1$ (red) and $G_2$ (blue) are shown. AE and TE represent the arrival and transmission of events, respectively. Initially, the channel is occupied (black region). So the arriving event of class $G_{1}$ has to be queued. Accordingly, its credit starts increasing. Once the channel is free, the packet corresponding to $G_{1}$ gets transmitted, and its credit decreases with a slope. Similar dynamics is plotted for $G_{2}$. B) i) The $\pi^2_{K}$ neuron integrates the first K temporal input events and generates a single output event based on a time-to-first-spike encoding scheme.  ii) As input events arrive at times $T_1$ to $T_4$, the membrane potential increases with a slope proportional to the number of inputs received. After receiving $K$ inputs, the slope remains constant. When the membrane potential crosses a predefined threshold $M$, the neuron emits an output event at $T$. The dynamics of a $\pi^2_{3}$ neuron is plotted here. iii) $\pi^2$-neuron - The figure illustrates the operation of the proposed CBS protocol, which aligns the credit-based traffic shaping (CBS) protocol with the dynamics of the $\pi^2_{K}$ neuron model. The modified credit dynamics of a $\pi^2_{2}$ neuron for the traffic classes $G_{1},G_{2}$ is highlighted here.  The bottom panels contrast traditional credit-based dynamics with the proposed $\pi^2_{K}$ dynamics. C) Traditional Asynchronous Traffic Shaping (ATS) mechanism: Incoming events are assigned to $2^P$ shaped queues based on their Priority Code Point (PCP) values. The ATS shaper delays each packet until its Transmission Eligibility Time (TET), calculated usually as a function of the traffic characteristics, is reached. Once eligible, events are forwarded to a shared queue (which is controlled by the modified CBS protocol to emulate the $\pi^2_{K}$ neuron dynamics) in a time-sorted order for transmission. The ATS protocol is re-envisioned to model the $\pi^2$ synapse. For the $i^{th}$ event, the $TET_{i}$ is computed as $T_{i} + W_{i}$, where $W_{i}$ is the synaptic delay (PCP code) associated with it. D) Illustration of a fully connected (n x n) $\pi^2$-NN architecture implemented using $\pi^2_{K}$ neurons, where synaptic weights are encoded as interconnect delays ($W_1$, $W_2$, $W_3$, \ldots).}
\label{fig2}
\end{figure*}
% MADHU FILL THIS SECTION BASED ON THE TEXT BELOW
% - First describe the k-neuron model
% - then describe what is credit-based shaping
% - then how CBS is modified to function as a k-neuron model
%\textbf{Re-interpreting the CBS protocol for neuron computation:} 
Ethernet switches are designed to handle a diverse mix of traffic types, ranging from time-sensitive audio/video streams to control messages and best-effort data. The switches also incorporate built-in mechanisms to support differentiated treatment of these different types of traffic. To manage this heterogeneity, switches implement multiple output queues at each port, with each queue corresponding to a specific traffic class or priority level. The flow of data through these queues is regulated by traffic shaping protocols, which control the timing and rate of packet transmission. A credit-based Shaping (CBS) protocol defined in the IEEE 802.1Qav standard regulates transmission by associating a dynamic credit counter with each queue \cite{cbs, nesting} as shown in Fig. \ref{fig2}A. A frame from a queue is eligible for transmission only when its corresponding credit is non-negative. Credits are updated based on the state of the queue: they accumulate at a fixed (idle) rate when frames are waiting, and decrease (send slope) during active transmission. This behavior is analogous to the integration and leakage dynamics in spiking neurons, where the membrane potential builds up before an event is generated and then resets or decays after the event.

We propose a modification of the CBS algorithm to implement a $\pi^2_{K}$ neuron model. Specifically, the $\pi^2_{K}$ neuron model uses a variant of the time-domain margin-propagation (TEMP) algorithm \cite{temp} and in the Methods section~\ref{secM2} we describe their equivalence. The $\pi^2_{K}$ neuron generates an event at time $T$, that is computed as follows;
\begin{equation}
T = \frac{M}{K} + \frac{1}{K} \sum_{j=1}^{K} t_{[j]}, \quad \text{where } t_{[j]} \in \{t_1, t_2, \ldots, t_d\}
\label{k11}
\end{equation}
Let $\{t_1, t_2, \ldots, t_d\}$ denote the set of input event arrival times, where $d$ is the total number of inputs received by the neuron. The terms $t_{[1]}, t_{[2]}, \ldots, t_{[K]}$ represent the first $K$ \emph{earliest} arrival times among the inputs. Thus, the neuron computes the mean of the earliest $K$ inputs and adds a constant offset $M/K$, resulting in the output time $T$. The constant $M/K$ ensures causality such that $T \geq t_{[K]}$. In a CBS-based model, credit accumulation mirrors the neuron’s membrane potential dynamics, with the number of incoming events modulating the accumulation rate (slope), which is represented here by the number of frames buffered in a shared queue. When the accumulated credit crosses a predefined threshold (M), the queue gate is enabled, and the buffered events are transmitted, analogous to a neuron’s firing event. Also, every shared queue is sized to accommodate only $K$ frames/events. The remaining incoming events are automatically dropped as portrayed in Fig. \ref{fig2}A. If the threshold is not reached before a pre-defined time-out value $T_{out}$, then all the events in the shared queue are dropped, and the credits are reset. The dynamics of the $\pi^2_{K}$ neuron is depicted in Fig. \ref{fig2}B. { The $\pi^2$ computation requires modest microarchitectural extensions to the CBS implementation—specifically programmable credit thresholds (an extension discussed in \cite{cbs_threshold} ) and queue length-dependent credit accumulation—which extends the internal credit-update logic (implemented using a $\log_2(K)$-bit counter shown in Fig. \ref{cbs} in the Supplementary section \ref{secA31}). Additional details of the modifications required in the conventional CBS algorithm to emulate the $\pi^2$ dynamics are presented in the supplementary section \ref{secA31}(Fig. \ref{cbs} and Table \ref{t1}). Note that these changes are only applied to the microarchitecture of the CBS shaper; the Ethernet frame formats are still compliant with the IEEE standard.}

%\newline
%\textbf{$\pi^2$-CBS}: 

\subsection{Asynchronous traffic shaping act as $\pi^2$ synapses}
% MADHU FILL THIS SECTION BASED ON THE TEXT BELOW
% - First describe the Pi2 synaptic model - using delays and routing
% - then describe what is ATS
% - then show ATS can be used as programmable synapses
\begin{figure*}[!htbp]
\centering
\includegraphics[width=1\textwidth]{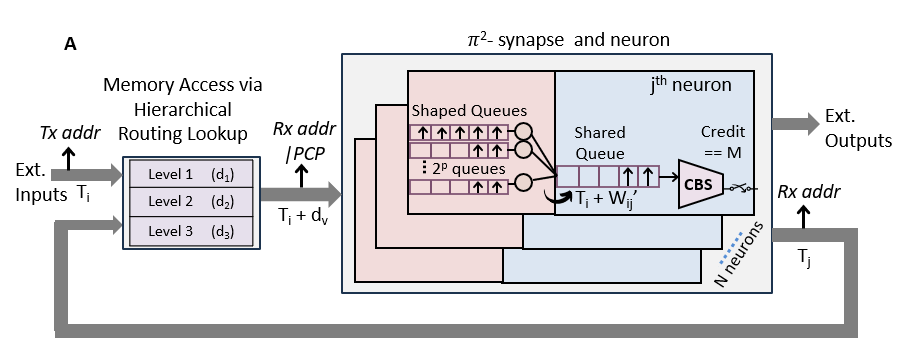}
\caption{ A) Realization of a $\pi^2$-NN architecture using a hierarchical routing table and modified traffic shaping protocols.
Incoming events—e.g., an event generated by the $i^{\text{th}}$ transmitter neuron at time $T_i$—are tagged with their source address and routed to the appropriate destination using a hierarchical routing lookup. The routing table resolves the destination address and assigns a $p$-bit Priority Code Point (PCP) value that encodes the synaptic delay. For a connection to the $j^{\text{th}}$ destination neuron, this delay is denoted as $W_{ij}$. The synaptic delay $W_{ij}$ is composed of two components: a memory access delay ($d_v$ - delay for the $v^{th}$ value), determined by the depth of the hierarchical routing structure, and a queuing delay ($W_{ij}'$), enforced by the ATS protocol. Once the eligibility time is reached, the ATS protocol forwards the event to a shared queue governed by the modified CBS protocol, as described in Fig.~\ref{fig2}C.}
\label{fig2B}
\end{figure*}
The Asynchronous Traffic Shaping (ATS) protocol is another traffic shaping algorithm defined in the IEEE 802.1Qcr~\cite{tsnintro}, designed for regulating the transmission of frames in Ethernet switches—particularly in time-sensitive networks. Unlike time-aware shapers that rely on global time synchronization, ATS operates asynchronously by assigning a Transmission Eligibility Time (TET) to each frame buffered in the output queues \cite{tsnintro} (Fig.~\ref{fig2}C). The TET determines when a frame becomes eligible for transmission, based on local parameters such as frame arrival time, burst, and frame size. In this regard, the ATS protocol facilitates precise temporal coordination without relying on centralized synchronization. Each synaptic delay is quantized and encoded into a $p$-bit Priority Code Point (PCP)~\cite{nesting}. The PCP field, defined in the IEEE 802.1Q VLAN tag (usually a 3-bit field), enables traffic differentiation by assigning each frame a priority level. Ethernet switches utilize this field to select the appropriate queue, apply shaping policies, and prioritize time-sensitive traffic. To enable delay-based event communication within the $\pi^2$ network, the modified ATS protocol supports three core neuromorphic operations:
\begin{itemize}
    \item {\bf Sorting:} Incoming events directed to a neuron are buffered into separate shaped queues based on their quantized synaptic delay, as specified by the $p$-bit PCP. This ensures temporal sorting within each delay bin, with events ordered by their arrival times. A total of $2^p$ shaped queues are instantiated per neuron, each corresponding to a specific quantized delay level.
    \item {\bf Delaying:} Let $T_i$ represent the arrival time of the $i^\text{th}$ frame and $W_i$ its corresponding synaptic delay. The Transmission Eligibility Time (TET) assigned by the modified ATS shaper is computed as:
    \begin{equation}
    TET = T_i + W_i
    \label{eq10}
    \end{equation}
    Each frame remains buffered in its designated queue until the system time reaches its computed TET, at which point it becomes eligible for transmission to the shared queue. As events are released based on TET, they are naturally inserted into the shared queue in a time-sorted manner.
    \item {\bf Event Dropping:} To limit queue congestion, each shaped queue is provisioned with a finite buffer capacity capable of storing up to $K$ events. Any additional frame arriving after the buffer is full is dropped, enforcing a form of traffic policing.
\end{itemize}
We extend the Asynchronous Traffic Shaping (ATS) algorithm to compute Transmission Eligibility Time (TET) values based on the synaptic delay parameters in a $\pi^2$ neural network. {The modification of the ATS protocol (IEEE 802.1Qcr) to support $\pi^2$ computation is presented in the supplementary section \ref{secA31} (Algorithm \ref{alg:ats}). Note that the modification does not add any new variables or extra computation; it simply repurposes a precomputed value to introduce delays.} An example of a fully connected ($N \times N$) instantiation of a $\pi^2$ neural network is shown in Fig.\ref{fig2}D. Let $T_i$ denote the event generation time of the $i^\text{th}$ presynaptic (Tx) neuron, and let $W_{ij}$ represent the synaptic delay from Tx neuron $i$ to postsynaptic (Rx) neuron $j$. The corresponding arrival time $\tau_{ij}$ of the event from neuron $i$ at neuron $j$ is given by:
\begin{equation}
\tau_{ij} = T_i + W_{ij}, \quad \forall i = 1, \dots, N
\end{equation}

In any processor architecture, delays are an unavoidable consequence of architectural features such as memory access latency, interconnect traversal time, and scheduling overhead. Rather than treating these delays as inefficiencies, the $\pi^2$ framework reinterprets them as functional components of synaptic processing. Fig.~\ref{fig2B}A illustrates a hardware-aligned approach on how these intrinsic delays are exploited through a combination of hierarchical routing and modified traffic shaping protocols. The synaptic delay can be decomposed as a sum of the memory access delay ($d_v$), determined by the depth of the routing structure, and a queuing delay ($W_{ij}'$), determined by the ATS protocol. The delayed event is then processed by the modified CBS protocol, which emulates the integration and thresholding behavior of the $\pi^2_K$ neuron. Once an output event is generated, it is fed back through the same process of hierarchical routing and traffic shaping pipeline. The routing table determines the overall network topology (feedforward, recurrent, or hybrid) and the recursive event-driven computation with memory-access delay and delay-based scheduling proceeds till the completion of the inference task. For instance, in a feedforward network, the total recursion time is determined by the depth of the network.

\subsection{Mathematical model of a $\pi^2$ neuromorphic architecture }
% MADHU FILL THIS SECTION BASED ON THE TEXT BELOW
% - First describe how CBS and ATS is already implemented in current and future ethernet switches.
% - Describe the architecture of the switch (with buffers etc) and how it is mapped to a standard neural network.
% - Describe how this can be scaled up to brain-scale - by cascading 
 
%The main contribution of this paper is to show the feasibility of using existing interconnect fabrics (switches and routers) and the supporting network protocols to implement scalable $\pi^2$ neural networks. 
%the first demonstration of the feasibility of mapping $\pi^2$-NN onto commercially available, optimized switching hardware fabrics such as Ethernet switches and routers—core components of today’s data center networks (Fig. \ref{fig2}F). 

% \newline
% \newline
Here we present a mathematical model to describe the $\pi^2$ neuromorphic architecture. For the sake of simplicity, consider a feedforward multilayer perceptron (MLP) network comprising $N$ layers, each consisting of a linear transformation followed by a nonlinear activation. The network input is denoted by $X$, and the output $y$ is computed according to
\begin{equation} \label{eq:feedforward}
    y = W^{(N-1)} \odot \sigma\left(W^{(N-2)} \odot \cdots \odot \sigma\left(W^{(1)} \odot \sigma\left(W^{(0)} X\right)\right) \cdots \right)
\end{equation}
where $W^{n}$ is the synaptic weight matrix connecting layer ${n-1}$ to layer $n$, $\sigma(\cdot)$ denotes the activation function applied at the output of every layer, and $\odot$ represents the $\pi^2$ operations based on the CBS and ATS primitives described in Fig.~\ref{fig2}. Note that for a conventional deep neural network, the $\odot$ in equation~\ref{eq:feedforward} is based on the standard multiply-and-accumulate (MAC) operations interleaved with nonlinear transformations.
\newline
%For $\pi^2$ neural networks architecture can be constructed using $\pi^2_{K}$ neurons. In contrast to conventional neural networks that rely on multiply-and-accumulate operations, 
$\pi^2$-NN uses synaptic weights to introduce programmable delays in the arrival times of input events. A differential margin-propagation proposed in \cite{temp},\cite{mulp_mp} is used to approximate inner products and is applied to Eq. \ref{k11}. Accordingly, the weights and inputs of the $\pi^2$ network are represented in a differential formulation. The real-values d-dimensional input  X is encoded into differential event spike times and fed as input to $\pi^2-NN$ as follows:  
\begin{equation}
  T^{0+} = [A+X]_+, T^{0-}=[A-X]_+  
  \label{k88}
\end{equation}
 The synaptic weights (W) are encoded as differential synaptic delays represented as $W^+ = [B + W]_+,W^- = [B-W]_+$ for all layers. Here, A and B are arbitrary constants, and $[.]_+ = \max(0,.)$ is a ReLU operation that ensures that the information represented in time is non-negative. 
\newline
 The output of a $j^{th}$ neuron in a $\pi^2$-NN is two spikes/events denoted by their respective time of occurrences $T_{j}^{+}$ and $T_{j}^{-}$. These occurrences are computed as follows:
\begin{equation}
\tilde T^{+}_{j} = \frac{M}{K} + \frac{1}{K} \sum_{k=1}^{K} t_{[k],j}^{+} \   \text{and} \ \  \tilde T^{-}_{j} = \frac{M}{K} +  \frac{1}{K} \sum_{k=1}^{K} t_{[k],j}^{-} \quad \text{where} 
\label{k44}
\end{equation}
\begin{equation}
t_{[k],j}^{+} \in 
\left\{ T_i^{+} + W_{ij}^{+} \,\middle|\, i = 1, \ldots, d \right\} 
\cup 
\left\{ T_i^{-} + W_{ij}^{-} \,\middle|\, i = 1, \ldots, d \right\}
\label{k441}
\end{equation}
\begin{equation}
t_{[k],j}^{-} \in 
\left\{ T_i^{-} + W_{ij}^{+} \,\middle|\, i = 1, \ldots, d \right\} 
\cup 
\left\{ T_i^{+} + W_{ij}^{-} \,\middle|\, i = 1, \ldots, d \right\}
\label{k442}
\end{equation}
The terms $t_{[1]}, t_{[2]}, \ldots, t_{[K]}$ represent the first $K$ \emph{earliest} arrival times among the inputs defined in the sets. As described in \cite{mulp_mp}, the difference between the positive and negative differential spike times provides an approximation of a dot product operation. To obtain a single-ended output, the differential spike times are combined as follows, and a nonlinear activation function (ReLu) $\sigma(.)$ is subsequently applied:
{
\begin{equation}
    T_{j} =  \sigma\  (\alpha(\tilde T^{-}_{j} - \tilde T^{+}_{j}))
    \label{k77}
\end{equation}
\begin{equation}
     T^{+}_j = V + T_j, T^{-}_j= V -T_j
    \label{k99}
\end{equation}}
Here, $\alpha$ is a hyperparameter that helps in faster convergence during training. {The single-ended output in Eq.\ref{k77} is subsequently re-encoded into a differential form (Eq. \ref{k99}), consistent with the input encoding described in Eq.\ref{k88}, and propagated to the next layer for further processing. $V$ is a constant chosen such that causality in the time domain is maintained between the input and output events.} Such a mapping enables the embedding of event-driven computation within conventional Ethernet-based communication protocols. This approach opens the door to leveraging ongoing advancements in switching and routing hardware to efficiently scale neuromorphic computing architectures. Further details on the mathematical framework required for training $\pi^2$ neural networks using gradient-based methods are presented in the supplementary section \ref{seca21},\ref{seca22}.

\section{Results}\label{sec2}
% AFTER THIS THE SECTIONS SHOULD BE ORGANIZED AS:
% - RESULTS
%     - Results for a small-scale network (understanding how pi2 could be training and the relationship between pi2 parameters ... omnet simulator - show can be mapped onto real hardware, jitter analysis, extracting the model for large-scale and faster simulations)
%     - Results for a mid-scale network (understand the scaling of pi2 training ...)
% - DISCUSSIONS
%     - also add something about using congestion control algorithms and quality-of-service protocols directly for training pi2 networks - similarity with the loss-functions
% - METHODS
% NO CONCLUSIONS
\begin{figure*}[!htbp]
\centering
\includegraphics[width=\textwidth]{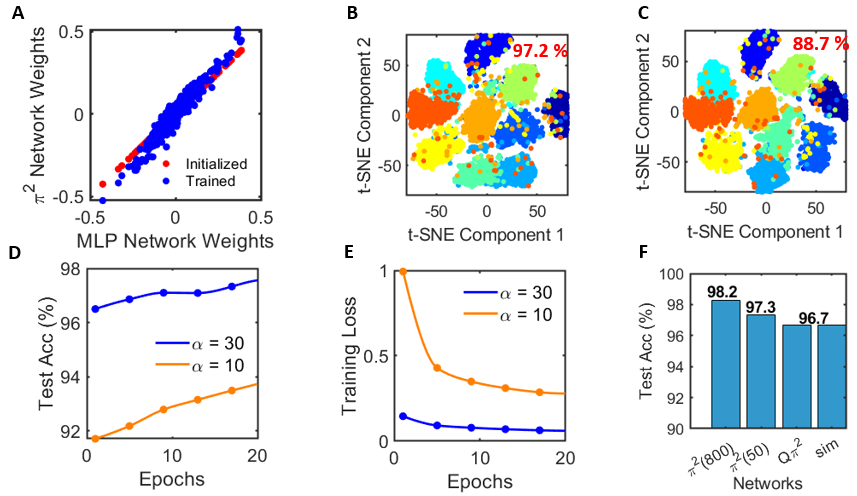}
\caption{A) Scatter plot portraying the relationship between weights of the $\pi^2$ and MLP networks (784x50x10 architecture) trained on the MNIST dataset. We chose this network due to the memory constraints of running the simulator on our system. When the trained MLP weights are directly mapped to the $\pi^2$ network weights, there is a 1-1 relationship. After retraining the $\pi^2$ network, a high correlation still exists. B) t-SNE plot of the hidden layer representations of the ten classes extracted from the trained MAC-based MLP network. Distinct clusters are obtained from different classes of the trained network. C) When the $\pi^2$ network is initialized with the trained MLP weights, the separation between the clusters still exists as it tries to approximate the MAC operation. However, an approximation error exists. This error propagates through the next consecutive layers, making the classification accuracy drop from 97.2\% to 88.7\%.  Each color represents a distinct class in these plots. D,E) The $\pi^2$ network has to be trained further to compensate for the drop in classification accuracy. The effect of the scaling parameter $\alpha$ on the test accuracy and training loss is portrayed here. Higher $\alpha$ helps in improving the separation between the classes to achieve robust classification. F) The 3-layer $\pi^2$ network with 50 nodes ($\pi^2(50)$) is retrained to achieve a classification accuracy of 97.34\%. The weights of the trained $\pi^2$ network are quantized to 3 bits ($Q\pi^2$) and simulated using Ethernet switches on OMNET++ software. The simulator supports 3-bit PCP codes and a network with 50 nodes, so we simulate accordingly. The simulation results (sim) exactly match the quantized networks' outputs to achieve a test accuracy of 96.67\%. By increasing the number of hidden nodes to 800, we can achieve the baseline accuracy reported in \cite{stanojevic2024} ($\pi^2(800)$).  }
\label{fig3}
\end{figure*}

\begin{figure*}[!htbp]
\centering
\includegraphics[width=\textwidth]{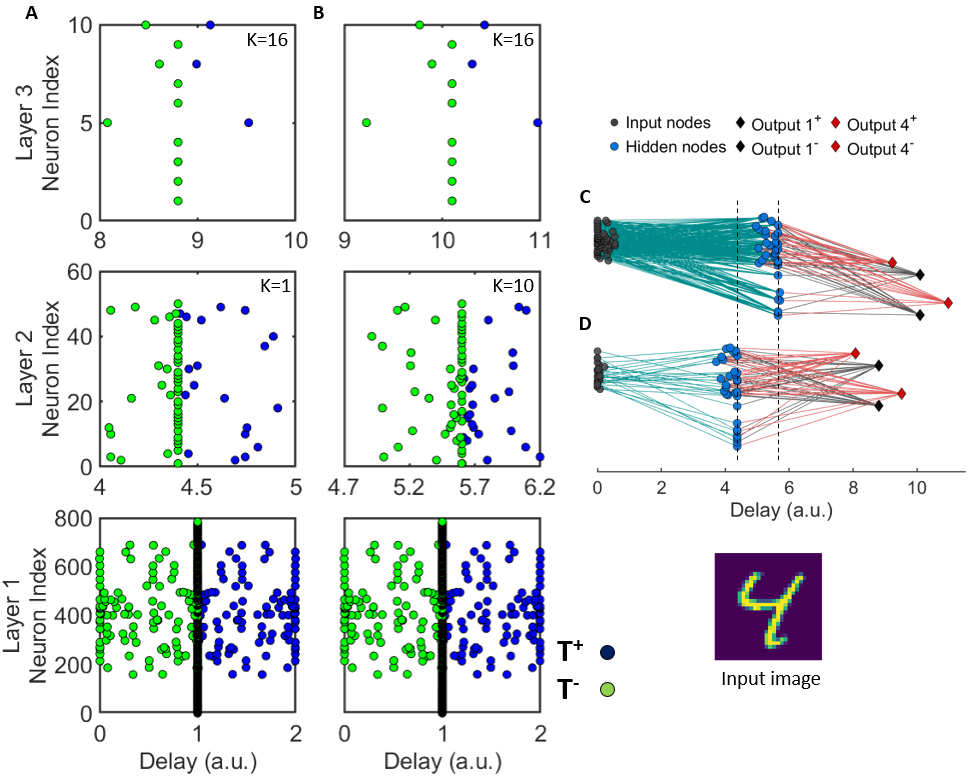}
\caption{A,B) The spatio-temporal patterns emerging from the layers (input (Layer 1), hidden (Layer 2), and output layer (Layer 3)) of the trained $\pi^2$ network are illustrated as a raster plot for different configurations of K. The blue and green dots represent the differential event generation times of the nodes in a layer. The population response to the input digit “4” is shown for (A) $K=[1,16]$ and (B) $K=[10,16]$ for the hidden and output layers, respectively. C,D) The corresponding evolution of neuronal activity for the same input is depicted for (C) $K=[1,16]$ and (D) $K=[10,16]$ (hidden and output layers). It can be observed that reducing the value of K to 1 leads to a computationally sparse (in terms of neuron and synaptic activity) and faster computation. Differential event activity (based on Eq. \ref{k88},\ref{k77} is selectively traced for classes “4” and “1” in these plots.}
\label{fig4}
\end{figure*}
%\textbf{Establishing a Mapping between Conventional MAC-Based Neural Networks and $\pi^2$-NNs:}  
%\textbf{Empirical Analysis:} To 
We first explore the functional equivalence between conventional neural networks using (multiply-accumulate, and non-linear activation operations) and $\pi^2$-NNs for a small-scale problem where we can analyze the intrinsic dynamics and learning in more detail. We consider a standard 3-layer multi-layer perceptron (MLP) with a 784-50-10 architecture and ReLU activation at the hidden layer, trained on the MNIST dataset. This particular configuration was selected because it could be executed using the OMNeT++ network simulator~\cite{omnet1}. The OMNET++ simulator is widely used in the networking literature~\cite{omnet_tsn,nesting,omnet2,atscbs2,fault} to benchmark different packet-routing and congestion control algorithms. As a result, the OMNET++ simulator supports many of the existing Ethernet protocol standards. { Several studies show that OMNeT++/INET can be aligned closely with real TSN hardware behavior \cite{tsn_hw_1,tsn_hw_2}. These studies demonstrate that the traffic-shaping primitives, such as the CBS and Time-aware shaping (TAS) protocols, can be reliably assessed in simulation and that the differences between simulated and hardware behavior are well-characterized and quantifiable.} However, due to the simulator memory constraints, this network architecture was chosen to simulate a model with fewer than 1,000 nodes. A comparable or smaller scale node count is also adopted for scalability studies using the OMNET++ simulator in \cite{nesting,omn_sc1,omn_sc2}. For this architecture, the MLP achieves a test accuracy of 97.2\%. The learned weights are then directly mapped to an equivalent $\pi^2$-NN, enabling evaluation of the model’s performance in the time domain. For specific values of the scaling parameter $\alpha=[30,30]$ and $K=[140,16]$ for the hidden and output layer, respectively, the $\pi^2$-NN achieves a test accuracy of 88.7\%. Despite the performance gap, both models exhibit strong functional resemblance in their learned representations. This similarity is visualized in Fig.~\ref{fig3}A,B,C using t-SNE projections of the hidden-layer activations from the MLP and the corresponding $\pi^2$-NN. A formal derivation of their equivalence is provided in the Methods section. Furthermore, we show that retraining the $\pi^2$-NN can recover the performance of the original MLP, effectively closing the performance gap introduced by direct weight transfer (Fig. \ref{fig3}D,E). Finally, we note the critical role of the scaling parameter $\alpha$ in training. Larger values of $\alpha$ enhance class separability and accelerate convergence during training. Fig.\ref{fig3}D,E illustrates this behavior by comparing training trajectories for different values of $\alpha$.  Furthermore, in contrast to conventional neural networks, information in $\pi^2$-NNs is encoded spatiotemporally, leveraging both the spatial structure of the network and the temporal dynamics of event-driven computation. This encoding paradigm enables $\pi^2$-NNs to exhibit rich and diverse combinatorial representational capabilities, as discussed in \cite{poly}. The emergent spatiotemporal representations across successive layers of the $\pi^2$-NN and the impact of the parameter K on the computational sparsity and latency of the networks are illustrated in Fig.~\ref{fig4}. The spatio-temporal activity observed for inputs of different classes is presented in the supplementary section \ref{secA4}(Fig. \ref{figmnist}).
\newline
To demonstrate the $\pi^2$-NN architecture can be mapped onto current packet-switching protocols, we implemented this network using Ethernet switches on the OMNeT++ network simulation platform \cite{omnet_tsn,nesting}. The simulator implements the core traffic shaping protocols as discussed in the previous section. The simulation results exactly match the software outputs of the 3-bit quantized network, as shown in Fig. \ref{fig3}F. The spatio-temporal patterns (similar to Fig. \ref{fig4}) extracted from the simulator are presented in the supplementary section \ref{secA4} (Fig. \ref{fig:tsn_mnist}).
 Due to memory limitations, simulation results are presented only for a small 3-layer $\pi^2$-NN comprising slightly fewer than 1,000 nodes. Notably, the differential nature of the $\pi^2$ algorithm effectively doubles the number of traffic flows, further increasing the simulation complexity. However, the accuracy of the network can be improved by increasing the number of hidden nodes to 800 (Fig. \ref{fig3}F), and we can match the baseline accuracy reported in \cite{stanojevic2024}.  Fig. \ref{fig3}F highlights that a 3-bit quantized $\pi^2$-NN can be mapped to the simulator without any drop in accuracy. So, for larger-scale $\pi^2$-NNs, scalability analysis is performed in software, and the corresponding results are reported based solely on software-level simulations in the subsequent experiments. {Note that our OMNeT++ results should be viewed as functional validation of the time-domain dynamics of $\pi^{2}$ rather than hardware-level performance claims. }
 % Current TSN ASICs do not expose the full combination of ATS, CBS, and programmable queue-gating needed to realize $\pi^{2}$ at scale, making direct hardware deployment premature. Our GPU and OMNeT++ experiments, therefore, validate the correctness of repurposing the protocols.
\begin{figure*}[!htbp]
\centering
\includegraphics[width=\textwidth]{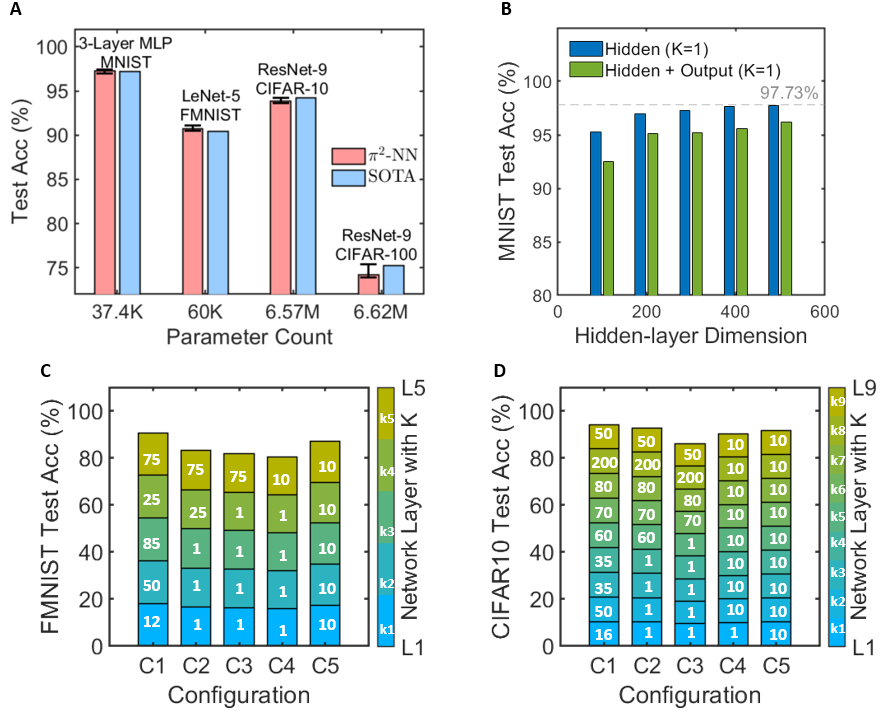}
\caption{Evaluation of $\pi^2$ Network Performance and Robustness. A) The test accuracy obtained on different $\pi^2$-NN architectures trained on standard image classification datasets - MNIST, FMNIST, CIFAR10, and CIFAR100 is reported. Average classification accuracy over 10 runs is reported here. 
B) Test accuracy obtained on a $\pi^2$ based 3-layer network trained on the MNIST dataset with varying hidden node dimensions. The blue bar represents the accuracy obtained when K is constrained to 1 at the hidden layer. The green bar represents the accuracy obtained when K is constrained to 1 at the hidden and output layers. C,D) Layer-wise configuration of $K$ in $\pi^2$ based LeNet-5 and ResNet9 models trained on FMNIST and CIFAR-10, respectively, showing the trade-off between computational sparsity and classification accuracy.}
\label{fig6}
\end{figure*}
\begin{figure*}[!htbp]
\centering
\includegraphics[width=\textwidth]{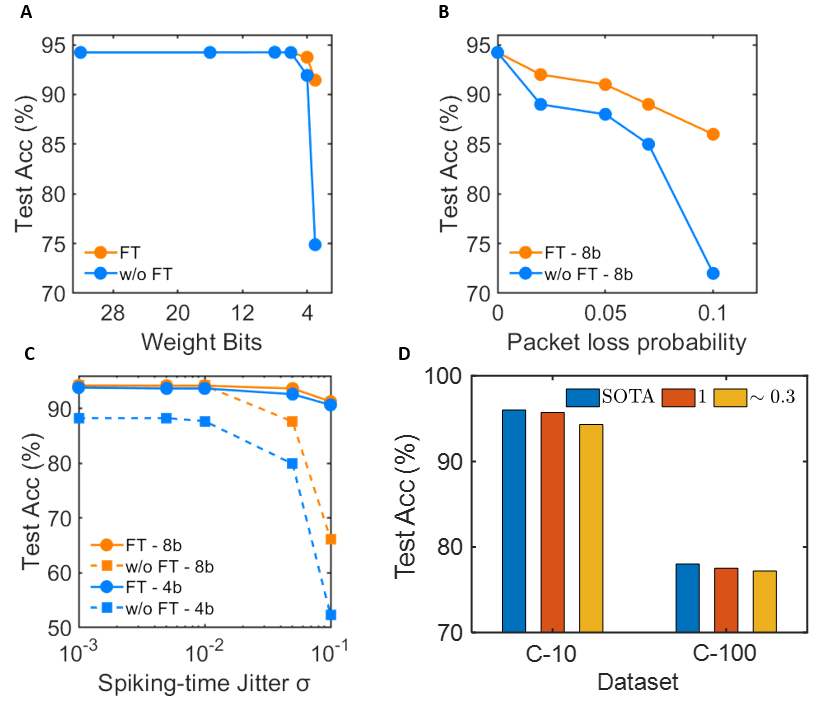} %FIG5_2.PNG
\caption{Evaluation of $\pi^2$ Network's Robustness. A)  Effect of weight quantization on accuracy with and without fine-tuning.  B) Accuracy as a function of the random packet/event drop probability applied independently at every layer. C) Accuracy as a function of the standard deviation (SD) of random noise values added to each spike time in the network (spiking time jitter). On fine-tuning for 10 epochs, the performance drop due to quantization, jitter, and random packet drops can be significantly improved. The average accuracy across 10 trials is reported here. D) {The test accuracy obtained using the $\pi^2_s$ algorithm on the CIFAR10 and CIFAR100 datasets. The accuracies for spiking sparsity ratios ($P/d$) 1 and 0.3 are reported and compared with the state-of-the-art \cite{r18}. The sparsity ratio is set to 1 for the input layer for all the experiments. }}
\label{fig61}
\end{figure*}
\newline
\newline
 We perform similar experiments on the F-MNIST dataset trained on the LeNet-5 network architecture comprising 60K parameters. We train the traditional MAC-based network and achieve an accuracy of 90.47\%. On porting the weights to an equivalent $\pi^2$-NN network with appropriately chosen values of $K$ and $\alpha$, we achieve an initial test accuracy of 85.23\%. 
Re-training the $\pi^2$-NN improves the accuracy to 90.7\%. Training a $\pi^2$-NN network from scratch with different random initializations leads to a 1\% accuracy drop on average. For comparison, \cite{stanojevic2024} reports a test accuracy of 90.94\% using a TTFS (Time-To-First-Spike) spiking neural network (SNN) variant of LeNet, while \cite{lenet} reports 90.1\% using an SNN-based LeNet model trained on the FMNIST dataset.
\newline
\newline
We extend the $\pi^2$-NN framework to larger-scale visual recognition tasks by training a $\pi^2$-based ResNet-9 architecture (comprising approximately 6.5 million parameters) on the CIFAR-10 and CIFAR-100 datasets. Directly initializing the $\pi^2$-based ResNet-9 with weights from its MAC-based counterpart results in negligible initial classification accuracy. We observe that as the network depth increases (i.e., more hidden layers are added), the test accuracy of the $\pi^2$-NN declines. This reduction correlates with a decrease in representational similarity between the two architectures, as measured by the correlation of hidden layer activations. These findings suggest that although $\pi^2$-NNs can approximate the function of MAC-based networks, the approximation error accumulates across layers, ultimately degrading both classification accuracy and feature alignment. 
\newline
\newline
Nevertheless, upon further training using the initialized weights, the $\pi^2$-ResNet-9 achieves a competitive maximum accuracy of 94.23\% on CIFAR-10. Prior works on ANN-based models report accuracies of 94.29\% and 90.8\% in \cite{cifar103} and \cite{cifar102}, respectively. A custom 9-layer residual convolutional network, DavidNet, achieves 94.08\% accuracy \cite{cifar101}, while a spiking ResNet-18 model (which has around 11.2M parameters) demonstrates 94.25\% accuracy in \cite{cifar104}. Notably, 94\% is also considered the benchmark for human-level performance on CIFAR-10 \cite{cifar105}. {We also experimented with training the network naively from scratch, without weight porting and knowledge distillation strategies, and achieved an accuracy of 89\%. }
\newline
\newline
For the CIFAR-100 dataset, the $\pi^2$-based ResNet-9 model achieves a maximum test accuracy of 75.5\%. In comparison, previous studies report accuracies of 75.41\% and 68.25\% on ResNet-9-based DNN models, in \cite{cifar1001} and \cite{cifar1002}, respectively. In \cite{cifar1003}, performance is further improved through model ensembling, resulting in an accuracy of 77\%. However, the accuracy achieved using a single ResNet-9 model in that study is 72.21\%. Further, researchers report an accuracy of 74.24 \% on ResNet-18-based SNNs in \cite{cifar104}. These comparative results are summarized in Fig.~\ref{fig6}A.  {Training the $\pi^2$-NN naively, from scratch, without weight porting and knowledge distillation strategies, resulted in an accuracy of 69\%. }Further details of the training configuration are provided in the Methods section~\ref{secM3}.
\newline
\newline
Training $\pi^2$-NNs on GPUs is both memory-intensive and computationally demanding due to their differential formulation and the nature of their core operations. While GPUs are highly optimized for parallel matrix–vector multiplications, $\pi^2$-NN training requires sorting and event-driven operations that are not inherently GPU-friendly \cite{causal}. As a result, we were unable to demonstrate results for significantly larger network architectures. All our experiments were conducted on a single 40GB NVIDIA A100 GPU. Training a $\pi^2$-ResNet-9 network on the CIFAR datasets with a batch size of 16 required approximately 34 GB of GPU memory, thus constraining the training of larger-scale networks. {It is important to note that the GPU experiments serve solely to verify the correctness of the learning rule and representational capacity of $\pi^2$-NN; the computational advantages emerge only when deployed on hardware where delay and sorting are physical rather than simulated operations. Thus, the results should not be interpreted as a demonstration of the computing in interconnects capability but rather as numerical emulations of the underlying dynamics of the time-domain equations. }
\newline
\newline
\textbf {Fine-tuning for hardware:}  One critical architectural lever for optimizing hardware efficiency is the choice of the parameter K. Reducing the value of K improves computational sparsity (as fewer inputs need to be processed, as portrayed in \ref{fig4}C,D), reduces hardware complexity, and minimizes the buffer and queue size requirements. Specifically, a simplified processor design can be implemented by setting K=1 in Eq. \ref{k11}. The $\pi^2$-Rx node generates an event immediately upon receiving the first input event, reducing the neuron model to a basic processing element—a switch. This kind of paradigm evaluates the interconnects' computational capabilities alone. While this approach significantly improves throughput by minimizing computational complexity to that of a switching operation (Methods section \ref{secM4}), it comes at the cost of reduced accuracy (Fig. \ref{fig6} B,C,D). Further, the $K=1$ networks also eliminate the memory access overhead of fetching the pre-computed values (membrane potential) from memory, thus enhancing the energy efficiency of $\pi^2$ networks \cite{1time_step,1time_step1}. However, it is important to note that the performance of $\pi^2$-NNs, as observed in software simulations, may degrade in real hardware due to physical limitations and non-idealities of the interconnection networks. Therefore, the network must be designed to be robust against hardware-level effects such as quantization, temporal noise (jitter), and packet drops.
\newline
\newline
{A related and equally important consideration for maintaining higher energy efficiency in neuromorphic systems is spiking sparsity. In the baseline $\pi^{2}$ formulation, each neuron emits two differential spikes/neuron (Eq. \ref{k88}-\ref{k77}), which increases event traffic. However, the fraction of spikes per neuron per input—the standard measure of communication sparsity—directly impacts energy consumption in neuromorphic systems \cite{stanojevic2024}. Sparse spiking is beneficial because (i) synaptic weight memory reads dominate energy in digital neuromorphic implementations, and unused synapses incur no cost, and (ii) spike-transmission cost becomes the dominant energy term as network size grows \cite{stanojevic2024}. Additionally, achieving and controlling high spiking sparsity is crucial for efficient bandwidth utilization.}
\newline
\newline
\textbf{Analyzing the tradeoffs between K and computational accuracy:}
We evaluate the computational capabilities of the interconnects using the $K=1$ paradigm within a 3-layer fully connected network trained on the MNIST dataset. Initially, $K$ is fixed at 1 in the hidden layer, while the output layer is allowed to use $K > 1$ during training. Subsequently, we constrain $K$ to 1 in the output layer as well, to assess the network's performance under fully constrained conditions. The results are summarized in Fig.~\ref{fig6}B. A baseline accuracy of 97.7\% was previously reported for a 3-layer MLP with 100 hidden neurons \cite{temp}. Applying the $K=1$ constraint in the hidden layer of the same architecture results in a 2\% drop in accuracy. However, increasing the hidden layer dimension to 500 restores the original performance. The trade-off in accuracy when K is constrained to 1 in the output layer as well is also reported. 
\newline
\newline
We extend a similar experiment to the $\pi^2$-NN-based LeNet-5 architecture trained on the FMNIST dataset. As illustrated in Fig.~\ref{fig6}C, imposing the \( K = 1 \) constraint on the first four layers of the network leads to an approximate 10\% reduction in test accuracy relative to the baseline. To mitigate this drop, we relax the constraint by setting \( K = 10 \) across all layers. This results in a more modest accuracy degradation of approximately 3\%, highlighting the trade-off between computational sparsity and model performance. { We evaluated simulation latency across different K configurations for a deeper LeNet-5 (five-layer CNN) trained on the FMNIST dataset and observed that the latency values span a very similar range across the tested settings. Specifically, when the configurations of K across the 5 layers were set to be - [12,50,85,25,75], [1,1,1,25,75], [10,10,10,10,10], we observed a simulation latency in the range of 30-35 time units. Reducing K to 1 did not lead to a significant reduction in latency. These results indicate that, while smaller K values can offer latency benefits in shallow networks, such benefits diminish in deeper architectures. Thus, K should be viewed as a design parameter that trades computational sparsity for accuracy, rather than as a knob that guarantees latency reduction at arbitrary depth.}
\newline
\newline
Setting $K=1$ for the first four convolutional layers of the ResNet-9 network trained on the CIFAR-10 dataset results in a modest accuracy drop of approximately 2\% relative to the baseline (Fig. \ref{fig6}D). However, extending this constraint to the first five layers leads to a significant decline in performance, with accuracy decreasing by nearly 10\%. To preserve accuracy, we relax the constraint and set $K=10$ on a few layers. This hybrid configuration limits the test accuracy loss to under 4\%. 91.5\% accuracy can be achieved if K is set to 10 across all layers. 
\newline
\newline
\textbf{Analyzing the effect of weight quantization, jitter, and packet drops:} The quantization of weight bits decides the number of shaped queues and PCP bits encoding required in the Ethernet-based implementation of $\pi^2$ networks (Fig. \ref{fig2}A,C). As shown in Fig. \ref{fig61}A, quantizing the ResNet-9 model on CIFAR-10 to 8-bit and 6-bit precision has a negligible effect on accuracy, maintaining baseline performance near 94.2\%. However, a significant degradation in accuracy is observed for a 3-bit network without fine-tuning. Notably, applying quantization-aware fine-tuning helps recover much of this loss for lower-precision networks. An accuracy above 90\% can be achieved using 3-bit quantized $\pi^2$ networks on fine-tuning. Inducing random packet drops in $\pi^2$-NN is similar to adding a dropout layer in conventional networks. Events are dropped independently in every layer with a predefined probability. We analyze the results of an 8-bit quantized ResNet-9 network trained on the CIFAR-10 dataset.   Spiking time jitter in $\pi^2$-NN is similar to activation noise in conventional deep neural networks \cite{stanojevic2024}. We investigate the results of an 8-bit and 4-bit quantized $\pi^2$-ResNet9 network trained on the CIFAR10 dataset. The results are reported in Fig. \ref{fig61}B,C, and the fine-tuning details are elaborated in the Methods section \ref{secM5},\ref{secM6}.
\newline
\newline
{\textbf{Analyzing the effects of spiking sparsity:}
Motivated by the energy benefits of spiking sparsity~\cite{stanojevic2024}, we introduce a 
sparse-spiking variant of $\pi^{2}$, denoted $\pi^{2}_s$, which reduces the spike 
count per neuron to fewer than one spike on average. From Eq.~\ref{k44}, the first 
$K$ earliest presynaptic events determine the postsynaptic spike times. For small 
$K$, many $T_i^{+}$ events contribute negligibly to $T_j^{+}$ and $T_j^{-}$, 
indicating that only a small subset of early arrivals meaningfully influences the 
computation.
Accordingly, in $\pi^{2}_s$ we remove the $T_i^{+}$ terms from the candidate sets 
in Eqs.~\ref{k441},\ref{k442}, and retain only the earliest $P$ events of $T_i^{-}$
($d \ge P \ge K$). The single-ended output in Eq.~\ref{k88}, previously encoded as 
a differential pair, is now represented by a single event occurring at $T_j^{-}$. 
Only these $P$ earliest events propagate to the next layer, yielding a spiking 
sparsity of $P/d$.
\newline
Mathematically, the real-valued $d$-dimensional input  X is encoded into a single  event spike time and fed as input to $\pi^2-NN$ as follows:  
\begin{equation}
   T^{0-}=[A-X]_+  
  \label{k88_1}
\end{equation}
 The synaptic weights (W) are encoded as differential synaptic delays represented as $W^+ = [B + W]_+,W^- = [B-W]_+$ for all layers. Here, A and B are arbitrary constants, and $[.]_+ = \max(0,.)$ is a ReLU operation that ensures that the information represented in time is non-negative. 
 The output of a $j^{th}$ neuron in a $\pi^2$-NN is one spike/event denoted by its respective time of occurrence $T_{j}^{-}$. This occurrence is computed as follows:
\begin{equation}
\tilde T^{+}_{j} = \frac{M}{K} + \frac{1}{K} \sum_{k=1}^{K} t_{[k],j}^{+} \   \text{and} \ \  \tilde T^{-}_{j} = \frac{M}{K} +  \frac{1}{K} \sum_{k=1}^{K} t_{[k],j}^{-} \quad \text{where} 
\label{k44_1}
\end{equation}
\begin{equation}
t_{[k],j}^{+} \in 
\left\{ T_{[i]}^{-} + W_{ij}^{-} \,\middle|\, i = 1, \ldots, P \right\}
\label{k441_1}
\end{equation}
\begin{equation}
t_{[k],j}^{-} \in  
\left\{ T_{[i]}^{-} + W_{ij}^{+} \,\middle|\, i = 1, \ldots, P \right\}
\label{k442_1}
\end{equation}
The terms $t_{[1]}, t_{[2]}, \ldots, t_{[K]}$ represent the first $K$ \emph{earliest} arrival times among the inputs defined in the sets. The term $T_{[i]}^-$ represents the $i^{th}$ earliest arriving presynaptic events. To obtain a single-ended output, the differential spike times are combined as follows, and a nonlinear activation function (ReLu) $\sigma(.)$ is subsequently applied:
\begin{equation}
    T_{j} =  \sigma\  (\alpha(\tilde T^{-}_{j} - \tilde T^{+}_{j}))
    \label{k77_1}
\end{equation}
\begin{equation}
    T_{j}^- =  V+ [A-T_{j}]_+ 
    \label{k77_1}
\end{equation}
Here, $\alpha$ is a hyperparameter, and V is a constant added to maintain causality. The post-synaptic event for the neuron $j$ occurs at a time $T_j^-$ and is propagated to the next layer for further processing.
This inherent selectivity provides a natural mechanism for enforcing high spiking sparsity, motivating the design of sparse-$\pi^{2}$ ($\pi^2_s$) variants that reduce event volume while maintaining functional performance.
\newline
When compared to the original $\pi^2$ formulation, the $\pi^2_s$ paradigm at least reduces the communication traffic by 50\%. Although the computation remains differential, only a single spike (the $T_j^{-}$ event) is transmitted forward, cutting communication traffic by half while preserving the temporal dynamics of the original $\pi^{2}$ formulation. The spiking sparsity reduces from 2 spikes/neuron in $\pi^2$-NN to $P/d$ ($\leq 1$) spikes/neuron. From the training perspective on GPUs, we can enable training of larger networks, as we now have to sort a much smaller set consisting of only P data points. Initially, we experimented with $(P/d)$ of 1 across all the layers of a ResNet-18 network and achieved an accuracy of 95.3\% and 77.2\% respectively on the CIFAR-10 and CIFAR-100 datasets. Further, inspired by the work in \cite{stanojevic2024} (they propose a spiking sparsity of 0.3 spikes per neuron), we experimented with a sparsity ratio of 1 in the first layer (as d was already small enough) and 0.3 in the remaining layers and achieved an accuracy of 95.2\% for CIFAR10 and 77.1\% for CIFAR100 datasets (0.7\% drop from SOTA \cite{r18}).   The results of the CIFAR datasets trained on the $\pi^2_s$ based ResNet-18-based network (11.2M parameters) with this algorithm are presented in Fig.\ref{fig61}d. The training details are elaborated in the Methods section \ref{secM7}.
\section{Discussion}

% \begin{figure*}[!htbp]
% \centering
% \includegraphics[width=0.7\textwidth]{Figures/FIG6.PNG}
% \caption{Trends in the performance projections for large-scale $\pi^2$ networks. }
% \label{fig8}
% \end{figure*}
%Recent deep-learning accelerators pursue energy-efficient inference, increasingly by harnessing physical substrates beyond conventional electronics \cite{pnn}. 
In this paper, we introduce a $\pi^2$ neuromorphic computing paradigm that leverages interconnect primitives: programmable delay, time-based sorting, and event dropping for computation and memory. Forward inference, therefore, becomes signal propagation and scheduling in time rather than the conventional multiply–accumulate (MAC) arithmetic. Viewed this way, $\pi^2$ offers a practical route to implement physical AI systems that are scalable while leveraging computational primitives inherent in the physical interconnect substrate (e.g., networking switches in packet-switching networks, waveguide delay lines in photonic networks\cite{photo}). 
%with learned functionality.
The key contribution of this work is to show how to re-purpose existing 
%Ethernet protocol primitives like 
% We strategize a mapping between the core operations of the $\pi^2$ paradigm and the existing capabilities of Ethernet protocols, as supported by modern Ethernet switches. Specifically, we reinterpret 
Ethernet traffic shaping protocols - namely the CBS and ATS protocols to model the $\pi^2_{K}$ neuron dynamics and synaptic operations. 
%To validate the feasibility of this approach, we implemented a hardware-aligned instantiation of a 3-layer $\pi^2$-NN on the OMNeT++ network simulation platform. The synaptic delays, quantized to 3 bits, were encoded using PCP tags within Ethernet frames, and neuron-level event propagation was emulated using ATS and CBS logic. 
By mapping neural operations directly onto these communication protocols, we transform commodity interconnect hardware into an in-network neuromorphic processor. To demonstrate this, we used an OMNET++ network simulator, where, due to the limitation of the simulator, we were only able to demonstrate a 3-layer $\pi^2$-NN network with 3-bit synaptic delays, which were encoded using PCP tags within the Ethernet frames. However, our modeling studies show that the $\pi^2$ paradigm can be scaled, with minimal to no drop in accuracy as the number of model parameters increases. {Notably, 3–4 bit effective precision is now consistent with frontier models, and the era of sub-2-bit LLMs underscores that meaningful computation does not require high numerical precision \cite{llm}. However, higher precision bits can be incorporated using different strategies - a) We can increase the bits associated with the PCP field of the VLAN tag (4 bytes long). This will increase the number of shaped queues required for ATS,  b) hierarchical delay decomposition, in which memory-access latency/routing delay provides coarse delay resolution while ATS-based queuing provides fine-grained scheduling, together enabling more than 8\,bits of effective precision (as presented in \ref{fig2B}); these approaches preserve throughput by retaining the pipeline structure of the $\pi^{2}$ neuron and synapse while allowing the effective synaptic precision to scale with minimal changes to the underlying switch microarchitecture. } {While validating these dynamics on hardware is ultimately desirable, access to commercial TSN platforms is limited and costly and lacks programmability, as documented in prior work \cite{tsn_hw_2}. Notably, delay-based computation is a common feature in hierarchical Address Event Representation (AER) implementations \cite{gert} and is supported on neuromorphic chips such as Loihi \cite{liohi_delay}, SpiNNaker \cite{spin_delay}, and TrueNorth \cite{TN_delay}. Temporal sorting, causality, and buffering are inherent features of AER, asynchronous NOCs \cite{temp,aer_sort}.  Furthermore, in such architectures, constraining the buffer size makes event dropping an inherent feature. Credit accumulation is synonymous with membrane potential integration in the existing spiking neuron models, such as the integrate-and-fire (IAF) neuron \cite{temp,iaf}.} However, the full scaling implications can only be realized when the $\pi^2$-NN is implemented on switching hardware, using commercial off-the-shelf (COTS) or custom chipsets, as we describe next.  
%Even though in this work, we were only able to demonstrate mapping limited by the sca 
%This fosters the growth of extensive network architectures while removing the need for extra hardware, significantly accelerating the transition from design to market compared to specialized processor solutions. 
%Furthermore, we evaluated the performance of $\pi^2$-NNs on standard image classification tasks, further reinforcing the viability of the $\pi^2$ paradigm as an efficient alternative to conventional neural processing. We empirically established functional equivalence between $\pi^2$ networks with the conventional MAC-based networks and characterized their information-processing behavior.

\begin{figure*}[!htbp]
\centering
\includegraphics[width=1\textwidth]{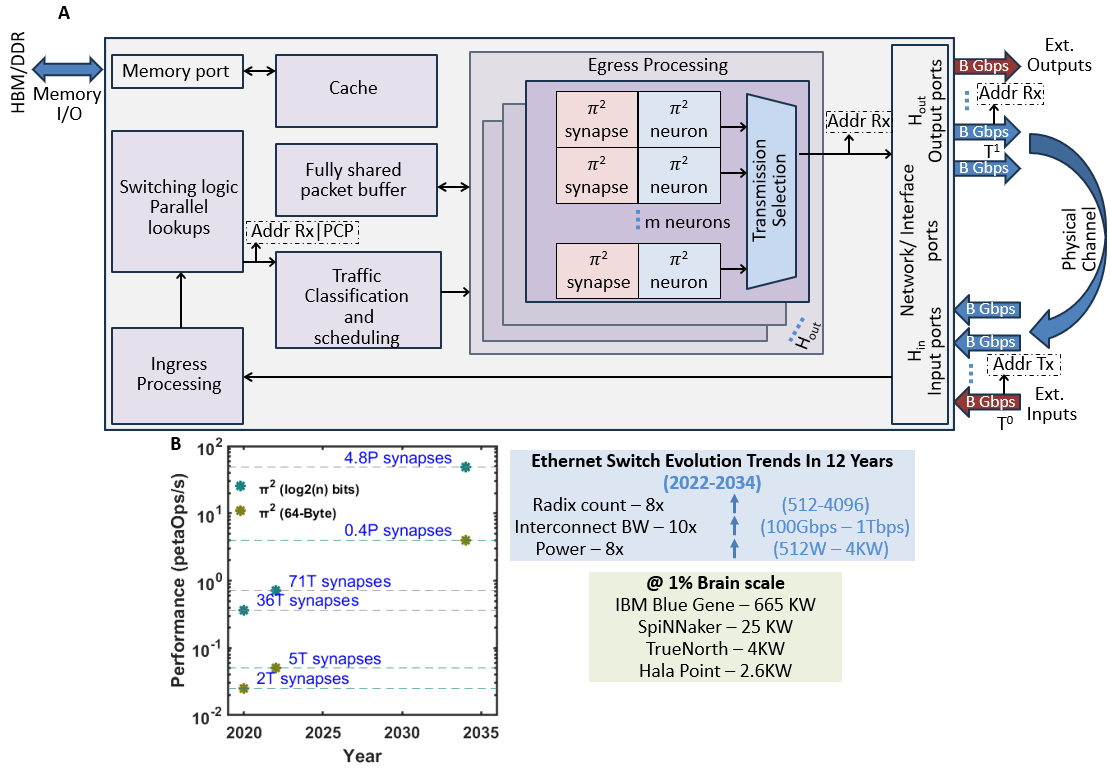}
\caption{ A) Architecture showing how the $\pi^2$-NN could be implemented on a high-speed packet-switching hardware platform. Each port of the switch supports a bandwidth of $B$ Gbps. Two ports are reserved for interfacing with external inputs and outputs, while the remaining ports are used for intra-core communication. Incoming events (represented by their corresponding source (Tx) addresses) are parsed and classified based on their priority (PCP code) and destination address (Rx address) by the ingress processing and traffic classification unit. The ATS mechanism is reinterpreted to emulate a $\pi^2$-synapse by introducing controlled delays, and the CBS mechanism is modified to replicate the spiking behavior of a $\pi^2_K$ neuron. These processing blocks can be cascaded within the switch to emulate $m$ neurons and their corresponding synapses at every output port. The output events—such as one generated at time $T^1$ from an input at $T^0$—can be recursively routed back to the input ports for continued processing. Here, the superscript $T^v$ denotes the processing order of the event. B){ Projected performance (petaOps/s) of $\pi^2$-based neuromorphic systems based on historical and extrapolated Ethernet switch trends, comparing two event-encoding granularities: (i) compact neuromorphic events consisting of $\log_2(n)$ bits, where 
 n is the number of neurons per core, and (ii) standard Ethernet frames with a minimum size of 64 B.}}
\label{fig7}
\end{figure*}

\textbf{Ethernet switches as $\pi^2$ neuromorphic processors :}
Modern switches now support the IEEE 802.1Q protocol suite, enable VLAN tagging \cite{ciscotsn}, and incorporate advanced packet processing capabilities alongside traffic management features~\cite{broadcomm12.8,25.6,marvel_tera}, including the CBS and ATS primitives summarized in Fig.~\ref{fig2}. %such as multiple queues per output port for granular flow control \cite{broadcomm12.8,25.6,marvel_tera}. 
Thus, an Ethernet switch/router~\cite{cpo, cisco, hpc} by itself can be used as an architecturally aligned substrate for the efficient and scalable realization of the $\pi^2$ computing paradigm. {Although minor, protocol-aligned extensions/modifications to existing traffic-shaping mechanisms are required, $\pi^2$ can benefit from ongoing advances in interconnect and networking hardware technologies}, as discussed in section~\ref{sec1}. Although the current standards do not yet support the combination of multiple traffic shapers on the same queue, extensive analysis of such ideas has been done in recent works \cite{atscbs1,atscbs2}, which we leverage here for the proposed $\pi^2$ architecture. {Further \cite{tsn_comb} proposes a traffic shaping engine (TSE) for next-generation TSN-compliant NoC/SoC platforms, with generic hardware explicitly supporting the selection of different traffic-shaping mechanisms.  Therefore, it can be anticipated that such capabilities will be available in the future.}
\newline
\newline
Figure~\ref{fig7}A shows the architecture of the $\pi^2$ neuromorphic core in which a population of neurons communicates over a high-performance Ethernet switching fabric. The switch is provisioned with $H_{in}$ ingress ports and $H_{out}$ egress ports. Among these, one input and one output port serve as interfaces to external systems, while the remaining ports facilitate intra-core communication. Each port is shared by $m$ neurons and operates with a bandwidth of $B$~Gbps, enabling high-throughput event exchange. The arriving frames, represented by their source addresses, are parsed and classified by the ingress processing and traffic classification unit. Within the egress pipeline, each output port is shared by a pool of $m$ neurons and their corresponding synapses, modeled using the proposed CBS and ATS traffic shaping protocols. These model the $\pi^2$ computing paradigm for event processing and communication. This pipeline is replicated in parallel across every output port to accommodate $n = H_{out}m$ neurons. The details of the implementation are elaborated in the supplementary section \ref{secA32} (Fig. \ref{fig10}).

{\textbf{Performance projections for large-scale $\pi^2$ architectures:}
The evolution of Ethernet switch ASICs has followed a sustained and favorable scaling trajectory, with switching capacity doubling approximately every two years due to alternating advances in SerDes lane count and per-lane signaling rate~\cite{cisco}. Over the past 12 years, switch radix, aggregate bandwidth, and total ASIC power have increased by approximately $8\times$, $10\times$, and $8\times$, respectively~\cite{cisco}. Extrapolating these trends suggests an additional $\sim 80\times$ increase in switch bandwidth over the next 12 years, enabling future-class switches with 4096 ports operating at 1~Tb/s per port at an estimated power of $\sim$4~kW (assuming $\sim$1~pJ/bit). In parallel, advances in co-packaged optics and 3D-stacked memory technologies have reduced interconnect and HBM energy efficiencies toward the sub-1~pJ/bit regime~\cite{cpoprogress,hbm_energy7,hbm_energy8}. Together, these trends point toward future systems dominated by bandwidth-rich, energy-efficient interconnect and memory fabrics.
\newline
Under the $\pi^2$ paradigm, these trends fundamentally alter the role of the interconnect: when computation is embedded directly within interconnect dynamics, the effective number of operations increases monotonically with available bandwidth (Fig.~\ref{fig7}B), converting communication energy—traditionally treated as overhead—into useful computation. The associated technology trends and projected performance scaling of $\pi^2$ networks are discussed further in the supplementary section \ref{secA32}, \ref{secA33} (Table \ref{table2}, \ref{table3},\ref{table3_e}). The performance projections presented here should be interpreted as architectural feasibility trends and idealized upper bounds for a {dedicated} $\pi^2$ deployment, in which a high-bandwidth switch datapath is reserved exclusively for $\pi^2$ event processing and integrated with external memory (HBM/DRAM). The analysis assumes sufficient routing capacity and memory bandwidth; in practice, memory capacity and access energy will impose additional constraints, particularly at brain scale. These limitations, however, are common to large-scale neuromorphic and HPC systems and are not unique to $\pi^2$. Accordingly, this work should be viewed as an architectural feasibility study rather than a claim about the capabilities of existing commercial Ethernet switches.
\newline
{A plausible approach to overcoming the limited on-chip memory of individual Ethernet switches is to organize a $\pi^2$ system as a distributed compute–memory mesh of interconnected switches. This design is conceptually related to hierarchical Address-Event Representation (AER) architectures, which achieve scalability by distributing routing and neural state across multiple stages rather than relying on centralized memory \cite{gert,aer_mem}. As in hierarchical AER, each switch maintains only local routing information and implements a limited subset of neurons and synapses. However, $\pi^2$ extends this model by explicitly repurposing routing latency as a computational primitive: events are forwarded using standard packet-switching mechanisms, while the number of hops traversed through the mesh encodes synaptic delay. In this way, network traversal directly implements neural timing rather than acting as communication overhead, enabling scalability by trading local memory capacity for network depth. This perspective aligns with the TEMP computational paradigm, which exploits intrinsic interconnect timing semantics for event-driven computation \cite{temp}, and suggests that interconnect hierarchy can serve as a viable substrate for large-scale neuromorphic computation.}
\newline
What distinguishes $\pi^2$ from conventional systems is that, in today’s interconnect-centric platforms, Ethernet switches minimally contribute towards the effective computation, despite consuming a substantial fraction of total system power. By contrast, $\pi^2$ repurposes this otherwise passive communication energy to realize computation directly through interconnect dynamics. As a result, even when constrained by realistic packet sizes and memory bottlenecks, the interconnect transitions from an energy sink to an active computational substrate. This shift fundamentally alters the performance–energy trade-off, suggesting that future gains in interconnect bandwidth—driven by advances in SerDes, radix, and co-packaged optics—can be directly converted into increased effective computation under the $\pi^2$ model.
\newline
The architectural stance adopted in this work assumes a dedicated Ethernet switch datapath for $\pi^2$ processing consistent with other neuromorphic platforms, such as Loihi \cite{liohi_delay} and Spinnaker\cite{spin_delay}, that also rely on dedicated interconnect and compute resources. In conventional neuromorphic substrates, scaling is limited by custom chip manufacturing and by challenges in memory–compute integration. However, this is not the case with Ethernet switches, whose capacity doubles every two years and is moving towards more energy-efficient processing. Thus, a dedicated switch deployment preserves the scalability advantages of networking hardware while enabling in-network computation. That said, multi-tenancy is feasible if required. Ethernet switches already support strong traffic-isolation mechanisms that separate best-effort and time-sensitive flows, including the use of distinct VLAN domains, reserved hardware queues \cite{queue}, and frame preemption strategies \cite{fp} to guarantee QoS. These mechanisms could be leveraged to isolate $\pi^{2}$ traffic from conventional workloads, though performance would then depend on the level of resource reservation provided to $\pi^{2}$.}
 \newline
\textbf{Spatiotemporal encoding in $\pi^2$-NN :}
In $\pi^2$-NN, the event encoding scheme is spatiotemporal; that is, the information is carried by the subset of neurons that activate in each layer and, within that subset, by their temporal order of spiking. Because synaptic delays allow a single neuron's event to be delayed by multiple offsets, a given unit can participate in multiple contributing groups that drive the next layer (supplementary section \ref{secA2} (Fig. \ref{xor})). Selectivity, therefore, arises from spatial sparsity (which units participate) and temporal ranking (when they arrive). This joint code is inherently combinatorial, as the number of distinct patterns grows with both the set of active neurons and their permutations in time \cite{poly}\cite{dendrocentric}. Further, as presented in Eq. \ref{k11}, every $\pi^2_{K}$ neuron is activated by the earliest arriving K events. The identity of the K earliest arriving spikes provides a natural back-pointer, enabling complete reconstruction of the causal route. This equivalence yields two interesting aspects: interpretability (explicit paths) and computational sparsity (only contenders on the best path are evaluated). From a training perspective, the back-pointers make credit assignment hardware-friendly as the synaptic weight updates are sparse (only K paths for every contributing neuron have to be updated), and the update rule is simple too \cite{mulp_mp}. Thus, $\pi^2$ networks are ideal for on-chip training strategies where memory, bandwidth, and energy are scarce. 
\newline
A single knob, K, sets how many earliest arrivals each neuron exploits, thus controlling computational sparsity and performance. Specifically, in K=1 networks, the first output to spike is the globally shortest path through the layered network; no analog accumulation is required. In this regime, the network follows shortest-delay paths, maximizing sparsity and minimizing latency. The predicted class is simply the output that first registers an event, making K=1 especially attractive for extreme sparse computing applications. {Also, the proposed formulation is closely related to established temporal coding strategies such as M-of-N, K-based, and rank-order spike encoding \cite{mn,dendrocentric,furber}. As noted in prior work \cite{temp}, these encoding schemes are well-suited to neuromorphic hardware, offer higher information content per spike, and promote energy-efficient sparse computation by emphasizing temporal order rather than firing rates \cite{dendrocentric,ttfs}.} Even though in this work, we have not fully exploited polychronicity to improve the capacity of the $\pi^2$-NN, future research will be able to use the proposed training infrastructure to incorporate such an encoding. In this work, we showed how $\pi^2$ neural networks can be efficiently trained using standard GPU infrastructure.  {Training—whether distilled, or native—is a one-time cost, whereas {inference is estimated to account for up to 90\% of the total computational cost of large-scale AI deployments \cite{aws}}. Thus, even if teacher networks are used during the training process, the long-term computational and energy benefits of $\pi^{2}$ arise entirely during inference, where the architecture operates independently of any conventional hardware substrate. } Further, the training time could be significantly reduced once dedicated $\pi^2$ hardware is available. This is because GPUs have been optimized to execute standard multiply-accumulate (MAC) and matrix operations, whereas the fundamental operations in a $\pi^2$ network are sorting and delays, which, as we showed in this paper, can be efficiently executed on a switching and interconnect hardware. {This bottleneck reflects a constraint of current hardware, not a limitation of the $\pi^{2}$ learning rule.} A dedicated $\pi^{2}$ accelerator would remove the GPU bottleneck entirely and enable large-scale training without relying on MAC-based networks. {Thus, the observed training bottleneck is a technology limitation of current GPUs, not a fundamental restriction of the $\pi^{2}$ learning rule}. Future work will therefore focus on realizing a dedicated $\pi^2$ computing platform using COTS hardware or dedicated ASIC chipsets.  
%\backmatter
%\newpage
\section{Methods}
\subsection{Computing effective energy utilization $\eta$}\label{secM1}
\begin{table}[h!]
\centering
\begin{tabular}{|c|c|c|c|c|c|c|}
\hline
Year & Tech (nm) & Compute EE & Switch EE & Interconnect EE  & Conv-$\eta$ & $\pi^2-\eta$  \\
\hline
2014 & 28   & 2.76    & 96.875  & 55           & 0.017848   & {0.59}  \\
2018 & 16   & 1.64    & 23.4375 & 30           & 0.029776   & {0.71}    \\
2022 & 7    & 0.75    & 10      & 6.25         & 0.044118   & {0.61}  \\ 
\hline
\end{tabular}
 \caption{Energy utilization trends over time for conventional and $\pi^2$ architectures across various technology nodes. Energy efficiency (EE), measured in picojoules per bit (pJ/bit), is used as the key performance metric. The temporal evolution of switch technology is derived from data reported in \cite{cpo}. Projected compute energy efficiency values are interpolated from datasets presented in \cite{compute_energy}. Switch energy efficiency figures are sourced from \cite{marvel}, \cite{25.6},  and \cite{edgecore_switch}, while interconnect energy efficiency data are drawn from \cite{qsfp}, \cite{cpo}, and \cite{broadcommcpo}.  }
\label{table:1}
\end{table}
In large-scale conventional (neuromorphic) architectures with passive interconnects, multiple compute units exchange information via Ethernet-based protocols and switches, as illustrated in \cite{truenorth_large_scale}. In contrast, large-scale $\pi^2$-NN systems can be realized using a single high-capacity Ethernet switch that supports millions of neurons.
For conventional architectures, the energy utilization metric, denoted as Conv-$\eta$, is defined as the ratio of compute energy to the total system energy, comprising compute, switch, and interconnect energy. Given that switch and interconnect components dominate the energy budget, Conv-$\eta$ approaches zero, indicating poor effective utilization. In $\pi^2$ architecture, the utilization metric $\pi^2$-$\eta$ is defined as the {ratio of useful switch and interconnect energy to the total energy consumed by the switch and interconnect. This formulation reflects the integrated nature of computation and communication in $\pi^2$ systems, where the switch itself performs both functions efficiently. We distinguish between the portion of switch energy that actively contributes to $\pi^{2}$ operations—denoted {$E^{sw}_{comp}$ and comprising delay scheduling, temporal sorting, queue-gating, credit accumulation, and event generation—and the remaining energy that does not participate in computation, including leakage, idle energy, initialization, clock trees, and interface overhead ($E^{sw}_{base} = E_{sw} - E^{sw}_{comp}$) and express $\pi^2-\eta$ as $\frac{E_{comp}^{sw} + E_{interconnect}}{E_{sw} + E_{interconnect}}$. The ratio of $E_{comp}^{sw}:E_{sw}$ is projected to be 0.36 in \cite{spower}. The interconnect delays can be characterized and utilized for computation as well}. By utilizing these physical effects of interconnects and switch operations for computation rather than treating them as purely communicational overhead, we can push the $\eta$ much closer to 1. 
\newline
The peak observed in $\pi^2-\eta$ in Table 1 arises because, between 2014 and 2018, interconnect energy did not scale at the same rate as switch core and compute energy, causing the interconnect component to dominate the energy budget (45\% drop in interconnect energy). By 2022, however, the introduction of co-packaged optics (CPO) significantly reduced interconnect energy (79\% drop in interconnect energy), leading to the subsequent drop in $\eta$. It is important to note that the energy-utilization analysis assumes a dedicated-mode deployment, in which the switch fabric is reserved exclusively for $\pi^{2}$ event processing and does not simultaneously forward conventional data traffic.}

\subsection{$\pi^2$ primitives and standard neural network operations} \label{secM2}
In this section, we first establish that the proposed $\pi^2_{K}$ neuron is a variant of the TEMP neuron model. Then we demonstrate how the $\pi^2_K$ neuron formulation mimics inner-product computation, which is the primitive underlying conventional neural network layers.
\newline
\newline
\textbf{$\pi^2_K$ neuron is a variant of the TEMP neuron model}: Let $\mathcal{U} = \{t_1, t_2, \ldots, t_d\}$ denote the set of input event arrival times. 
For simplicity, we assume that $t_1 \leq t_2 \leq \cdots \leq t_d$
 For a fixed parameter $\gamma \geq 0$, the TEMP neuron generates a spike/event at time T when the following condition is satisfied \cite{temp}:
\begin{equation}
\sum_{j=1}^d \max(0, T - t_j) = \gamma,
\label{temp}
\end{equation}
Consider the subset that contributes to the spike generation at time \(T\) as $\mathcal{S}(T) \;=\; \{\, t_j \in \mathcal{U} \;:\; t_j < T, j \leq d \,\}$ with cardinality \(s = |\mathcal{S}(T)|\).
If '$s$' spikes contribute to the membrane potential before the neuron spikes at T time units, then the closed form solution of Eq.\ref{temp} can be obtained as follows:
\begin{equation}
    T \;=\; \frac{\gamma}{s} \;+\; \frac{1}{s} \sum_{u \in \mathcal{S}(T)} u .
    \label{temp1}
\end{equation}
Equivalently, we define $T \;=\; TEMP(\mathcal{U}, \gamma)$,
where $TEMP(\cdot, \cdot)$ denotes the function that maps the set of arrival times and parameter $\gamma$ to the spike output time $T$ satisfying Eq.~\ref{temp}.
For a fixed parameter $\gamma$, the value of '$s$' keeps varying for the nodes in a layer, as pre-synaptic input spike times vary. We relax the constraints in Eq. \ref{temp} and propose a $\pi^2_{K}$ neuron model, similar to TEMP's neuron model, where the number of events contributing to the spike-out time $T$ is fixed by a parameter K.
\newline
\newline
For a fixed integer $K \leq d$, the $\pi^2_{K}$ neuron produces at most one output spike at time T computed as:
\begin{equation}
   T = \frac{M}{K} + \sum_{j=1}^{K}\frac{t_{j}}{K}  \label{k20}
\end{equation}
Here, M is a positive constant that plays the role of $\gamma$ in Eq.\ref{temp} provided it is chosen such that $t_{K} \leq T \leq t_{K+1}$. Equivalently, we define $T \;=\; \pi^2_{K}(\mathcal{U}, K)$,
where $\pi^2_{K}(\cdot,\cdot)$ denotes the function that maps the ordered set of arrivals 
$\mathcal{U}$ and the fixed parameter $K$ to the spike time $T$ defined in Eq.~\ref{k20}.  Note that, for a fixed value of K, the value of $\gamma$ in Eq.\ref{temp} will vary for different inputs. To extract the value of $\gamma$ for a given M and K such that $\pi^2_{K}(\mathcal{U}, K) = TEMP(\mathcal{U},\gamma)$, we can compute the event generation time $T$ from Eq.\ref{k20} and substitute it in Eq.\ref{temp}. 
\newline
From a software implementation standpoint, the TEMP formulation requires a full sorting operation with complexity $O(d \log d)$, whereas $\pi^2_{K}$ requires only a partial sort (\texttt{topK}) with $O(d \log K)$ cost. From a hardware design perspective, fixing the number of contributing inputs enables precise estimation of buffer sizes, thereby simplifying resource allocation and improving the scalability of large-scale neuromorphic implementations. In particular, when $K=1$, the neuron model reduces to a simple switching primitive, which can be directly realized as the operation of a single transistor, with minimal buffer requirements.
\newline
\newline
\textbf{$\pi^2_{K}$ neurons can be interpreted as mimicking inner-product computation on their inputs}:  
At the core of the TEMP neuron model lies \emph{margin propagation} (MP), a piecewise-linear approximate computing technique introduced in~\cite{bias,mulp_mp,temp}. In~\cite{mulp_mp}, MP was employed to approximate inner products between vectors, and the error characteristics of this approximation were analyzed and quantified. Following a similar reasoning, the TEMP neuron model can also be interpreted as an approximation to the inner product.  
\noindent  
Let $\mathbf{w}, \mathbf{x} \in \mathbb{R}^D$ be $D$-dimensional real vectors. For convenience, we define the sets  
\[
\mathcal{U}^{+} = \{\, t_i : t_i = x_i + w_i, \; i = 1,\ldots,D \,\}, 
\quad
\mathcal{U}^{-} = \{\, t_i : t_i = x_i - w_i, \; i = 1,\ldots,D \,\}.
\]  
Given a parameter $\gamma \ge 0$, the dot product can be approximated as
\begin{equation}
y \;=\; \mathbf{w}^{\!T}\mathbf{x}
\;\approx\; -TEMP\big([\mathcal{U}^{+}, -\mathcal{U}^{+}],\,\gamma\big)
\;+\; TEMP\big([\mathcal{U}^{-}, -\mathcal{U}^{-}],\,\gamma\big),
\label{eq:mp_temp_pairs}
\end{equation}
where $TEMP(\cdot, \gamma)$ denotes the operator defined in Eq.~\ref{temp}. Eq.~\ref{eq:mp_temp_pairs} highlights the differential formulation inherited from MP to approximate the inner product.
Since we have established the equivalence between TEMP and the proposed $\pi^2_{K}$ neuron, the same relationship can be expressed in terms of $\pi^2_{K}$ as  
\begin{equation}
y \;=\; \mathbf{w}^{\!T}\mathbf{x}
\;\approx\; -\pi^2_{K}\big([\mathcal{U}^{+}, -\mathcal{U}^{+}],\,K\big)
\;+\; \pi^2_{K}\big([\mathcal{U}^{-}, -\mathcal{U}^{-}],\,K\big).
\label{eq:mp_inner_product_pi2}
\end{equation}
where $\pi^2_{K}(\cdot, K)$ denotes the fixed-$K$ operator defined in Eq.~\ref{k20}. Furthermore, empirical results presented in Fig.~\ref{fig3}A, B, and C show that the $\pi^2_K$ neuron mimics dot product computation.

In Fig. \ref{fig3}B,C, we plot the t-SNE representations of the MAC-based MLP and $\pi^2$ networks to analyze the functional similarities between their internal representations. To achieve this, we utilize the built-in TSNE function from Python’s sklearn.manifold module to reduce the high-dimensional hidden layer activations to a 2-dimensional space. The t-SNE is configured with a perplexity of 30 and 1000 iterations to ensure a meaningful and well-converged low-dimensional embedding.
\newline

\subsection{Training of $\pi^2$ neural networks on GPU} \label{secM3}
We adopt a knowledge transfer-based training strategy for the $\pi^2$-NN, wherein a conventional MAC-based neural network (Teacher network) is first trained to achieve baseline accuracy on a given task. The trained model's parameters are then used to initialize the $\pi^2$-NN, which is subsequently fine-tuned to match the performance of the teacher network. For all software-based simulations, we refer to the conventional MAC-based network as the teacher network and its corresponding $\pi^2$-NN implementation as the student network. For training the student network, we employ a distillation loss that combines the conventional cross-entropy loss with a temperature-scaled Kullback-Leibler (KL) divergence between the soft output distributions of the teacher and student networks. At the input layer, to convert the pixel values (X) to differential spike timing (T), as specified in \cite{temp}, we use the following strategy.
\begin{equation}
    T_{in}^+ = |3 + X|_+ \ \ \&\ \ W^+ = |3+W|_+
    \label{t+}
\end{equation}
\begin{equation}
    T_{in}^- = |3-X|_+ \ \ \&\ \ W^- = |3-W|_+
    \label{t-}
\end{equation}
A similar differential strategy was used to represent the synaptic weights (W) of the network across all layers, as described by the above equations. $|.|$ represents the ReLU operation. 
\newline
\newline
%\textbf{Simulation details:}
%\newline
\textbf{Training with a 3-layer feedforward architecture on the MNIST dataset:} The MNIST dataset was preprocessed by normalizing pixel intensities to the range [0, 1]. The teacher network was trained using standard stochastic gradient descent with a learning rate of 0.1. The network was optimized to minimize cross-entropy loss and achieved a test accuracy of 97.22\% within 30 training epochs. The student $\pi^2$-NN was initialized with the weights learned by the teacher network. It was then retrained for 20 epochs using the same learning rate. The offset parameters $A,B$ used for differential encoding of inputs and weights were set to 3 and 0, respectively.  The $\pi^2$-NN was configured using hyperparameters $K = [140, 16]$ and $\alpha = [30, 30]$ for the hidden and output layers, respectively. The results are presented in Fig. \ref{fig3}A-F. The trained synaptic weights were quantized to 3 bits, and their effect on the inference accuracy is reported. The same training strategy was implemented for training the 3-layer network with 800 nodes. The hyperparameters K and $\alpha$ were set to $[115,70]$ and $[100,100]$ respectively.  All experiments were conducted using mixed precision training using 1 A100 GPU.
\newline
\newline
\textbf{Training LeNet-5 architecture on the FashionMNIST dataset:} The FMNIST dataset was preprocessed by scaling pixel intensities to the range \([0, 1]\). The teacher network---a conventional MAC-based architecture---was trained using the Adamax optimizer with a learning rate of \(5 \times 10^{-3}\), minimizing the cross-entropy loss over 30 training epochs. The student $\pi^2$-NN was initialized with the pretrained weights from the teacher model and subsequently fine-tuned for 20 epochs using the same optimizer and learning rate. Differential encoding of inputs and weights was performed using offset parameters \(A = 3\) and \(B = 3\), respectively. The $\pi^2$-NN was configured with temporal sparsity parameters \(K = [12,\ 50,\ 8525,\ 75]\) and scaling factors \(\alpha = [1,\ 10,\ 10,\ 10,\ 10]\) across its respective layers.
\newline
\newline
\textbf{Training ResNet-9 architecture on the CIFAR-10 dataset:} The CIFAR-10 dataset, comprising 50,000 training and 10,000 test images across 10 classes, was preprocessed by applying standard data augmentation techniques commonly used in image classification tasks. Each training image was randomly cropped to $32 \times 32$ pixels with a padding of 4 using reflection padding, followed by random horizontal flipping. The pixel values were then normalized using the dataset-specific channel-wise mean and standard deviation values: $(\mu_R, \mu_G, \mu_B) = (0.4914, 0.4822, 0.4465)$ and $(\sigma_R, \sigma_G, \sigma_B) = (0.2470, 0.2435, 0.2616)$. The teacher network was trained using the Adam optimizer with a learning rate of $1 \times 10^{-2}$, a weight decay of $1 \times 10^{-4}$, and a OneCycleLR scheduler to achieve the baseline performance. {We initially  trained the $\pi^2$-NN naively (with the same hyperparameters), from scratch, using backpropagation and achieved a 5\% drop from the SOTA accuracy. To improve accuracy, we implemented the KD strategy.}
To train the $\pi^2$-NN student model on CIFAR-10, we adopt a knowledge distillation strategy, initializing the network using a pre-trained MAC-based ResNet9 teacher model. The K values corresponding to the 9 layers of the network were as follows:$K = [16, 50, 35, 35, 60, 70, 80, 200, 50]$. The $\pi^2$ network was trained using the Adam optimizer with a learning rate of $1 \times 10^{-2}$, a weight decay of $1 \times 10^{-8}$, and a OneCycleLR scheduler for both the experiments. Training was conducted over 60 epochs using a batch size of 16.
\newline
\newline
\textbf{Training the ResNet-9 architecture on the CIFAR-100 dataset:} We train the teacher ResNet-9 network on the CIFAR-100 dataset, comprising 50,000 training and 10,000 test images across 100 classes. Image preprocessing includes random cropping with 4-pixel padding, horizontal flipping, and normalization using channel-wise mean and standard deviation values of (0.5071, 0.4867, 0.4408) and (0.2675, 0.2565, 0.2761), respectively. The model is trained using the Adam optimizer with a learning rate of $1 \times 10^{-2}$ and a weight decay of $1 \times 10^{-4}$ for 50 epochs. A OneCycleLR scheduler is employed to modulate the learning rate with a three-phase annealing schedule. The K values used to train the $\pi^2$-NN student model on CIFAR-100 were  [10, 50, 55, 65, 100, 100, 80, 200, 100]. The network was trained using the Adam optimizer with a learning rate of $1 \times 10^{-2}$, a weight decay of $1 \times 10^{-8}$, and a OneCycleLR scheduler. Training was conducted over 60 epochs using a batch size of 16. {We also trained the $\pi^2$-NN naively, from scratch, using backpropagation and achieved a 5\% drop from the SOTA accuracy, with the same training setup. To improve accuracy, we implemented the KD strategy.}

\subsection{OMNET++ implementation of the MNIST $\pi^2$ network}
The nodes/neurons are simulated as TSN devices, compatible with the traffic-shaping protocols. The synaptic delays are quantized  and encoded within the Priority Code Point (PCP) fields of the incoming Ethernet frames. The OMNeT++ simulator supports a three-bit Priority Code Point (PCP) that is part of the VLAN tag \cite{nesting}. So, the synaptic delays are quantized to 3 bits. The input events are first routed to an Ethernet switch implementing the $\pi^2$-ATS protocol. The TET values are computed based on the PCP value of the frame (synaptic delay), and the events are buffered until their TET is reached. The delayed events are then transmitted to another Ethernet switch implementing the $\pi^2$-CBS protocol to simulate the $\pi^2_{K}$ dynamics. The implementation details are presented in the supplementary section \ref{secA4}.

\subsection{Training $\pi^2$ networks with constraint on K} \label{secM5}
\textbf{MNIST and FMNIST}: For experiments in Fig.~\ref{fig6}B,C, we adopt a progressive, layer-wise training strategy to transition the network into a full \( K=1 \) configuration. In this approach, layers are sequentially constrained to \( K=1 \), followed by fine-tuning and evaluation at each stage. This iterative procedure is continued until all layers operate under the \( K=1 \) constraint. 
\newline
\textbf{CIFAR-10 with ResNet-9}: For experiments in Fig.~\ref{fig6}D, a similar layer-wise strategy is employed for the ResNet-9 architecture trained on the CIFAR-10 dataset. However, constraining the first four convolutional layers to \( K=1 \) results in a substantial performance drop, with accuracy degrading by over 25\%. To mitigate this, we adopt a hybrid configuration where \( K \) takes values from the set \(\{1, 10\}\) across different layers. This selective relaxation helps preserve accuracy while retaining the benefits of temporal sparsity. For these experiments, the network is initialized from a pre-trained baseline and fine-tuned under the \( K \)-based constraints for 30 epochs. All other hyperparameters are retained from the original training setup.

\subsection{Calibrating for finite-precision, jitter and packet drops} \label{secM6}
We implement two types of hardware constraints: (i) spiking time jitter and (ii) weight quantization, and (iii) random packet drops. 
To incorporate weight quantization (Fig. \ref{fig61}A), we employ quantization-aware training (QAT), allowing weights to be represented with reduced bit precision during inference. Specifically, we fine-tune the network for 10 epochs using quantized weights, starting from a pretrained model. To incorporate spiking time jitter (Fig. \ref{fig61}C), a random value of a Gaussian noise of a given standard deviation is added to each spiking time of the TTFS-network inputs and the outputs of each layer, similar to \cite{stanojevic2024,jitter1}.  We fine-tune the network for 10 epochs using jitter and quantized weights (8 bits and 4 bits), starting from the pre-trained model. For packet drops  (Fig. \ref{fig61}B), we stochastically suppress events within a layer with a preset drop probability and assess the resulting performance; this condition is likewise fine-tuned for 10 epochs from the pretrained initialization. Across all robustness experiments, we use an initial learning rate of 1e-3 and keep all other hyperparameters identical to the original training configuration.
{\subsection{Training $\pi^2_s$ networks} \label{secM7}
 We train the $\pi^2_s$- ResNet-18 network on the CIFAR-10 dataset following a similar strategy of weight porting and knowledge distillation. The model is trained using the Adam optimizer with a learning rate of $1 \times 10^{-2}$ and a weight decay of $1 \times 10^{-4}$ for 75 epochs. A OneCycleLR scheduler is employed to modulate the learning rate with a three-phase annealing schedule. The sparsity value was set to 1 for the first layer. We experiment with a sparsity value of 1 at the input layer and 0.3 for the remaining layers and report the results. Further details on the training setup and the K values for the 18-layer network can be obtained from our shared code repository. }

\subsection{Computational complexity of $\pi^2$ networks} \label{secM4}
Consider a fully connected n $\times$ n MAC layer, consisting of n input and output neurons. The number of addition and multiplication operations performed at every node of the output layer is 
\begin{equation}
    Ops_m = nC_m + (n-1)C_a
\end{equation}
where $C_m$ and $C_a$ represent the computational complexity of performing a multiplication and addition operation.
For an n x n $\pi^2$ network following the $\pi^2_{K}$ neuron paradigm, the number of addition operations performed at every node of the output layer depends on the value of K. The delaying, time sorting, and event dropping operations are inherently implemented using interconnects. The number of operations at the neuron level is
\begin{equation}
    Ops_k = 2KC_a
\end{equation}
A factor of 2 accounts for the differential computation. For K=1 networks, no compute logic is required. An event is generated as soon as the first input reaches the output node. Thus, the computational complexity is equivalent to a switching operation of a transistor. The number of operations at every node reduces to 
\begin{equation}
    Ops_1 = 2C_s
\end{equation}
In the hardware implementation of conventional networks, the interconnects do the work of transmitting data to and from memory, external interfaces, communicating modules, and between different submodules within the hardware substrate. The communication complexities and memory accesses are usually ignored while computing the computational complexity in conventional networks \cite{spkdriven}.  Accordingly, in both cases, we have only considered the complexity of neuron-level computations. To emphasize the hardware efficiency of this approach, it is noteworthy that a 16-bit full adder typically requires approximately 250–500 transistors \cite{add}, while a 16-bit multiplier demands over 4,000 transistors \cite{mul}. Furthermore, the energy cost of a floating-point multiplication can be up to four times that of a floating-point addition \cite{horowitz}, underscoring the significant hardware and energy advantages of the $\pi^2_{K}$ neuron.

\section*{Acknowledgements}
The authors would like to thank Haresh Dagale (Principal Research Scientist, Department of Electronic Systems Engineering, IISc) for his insights on communication network protocols. The authors would also like to acknowledge the joint IISc-WashU Memorandum of Understanding, to facilitate the collaboration between the two institutions. M.S.R., C.S.T., and S.C. participated in the Neuromorphic Integrated Circuits workgroup at the Telluride Neuromorphic and Cognition Engineering workshop in 2024, Colorado, USA, which was supported by the National Science Foundation grant ECCS:2332166. M.S.R., C.S.T., and S.C. also acknowledge the 2025 Bangalore Neuromorphic Engineering workshop (BNEW 2025), where some of the $\pi^2$-NN discussions occurred. S.C. would like to acknowledge the WashU McDonnell International Scholars Academy for supporting travel to BNEW 2025 at IISc. C.S.T. would like to acknowledge the Pratiksha Trust for funding support.
\section*{Author contributions}
M.S.R., C.S.T., and S.C. formulated the concept of $\pi^2$-NN and the use of interconnect primitives. M.S.R. performed all the simulations and benchmarking. C.B. contributed to improving the training of $\pi^2$-NN. All authors/co-authors contributed to the proofreading and writing of the manuscript.

\section*{Code availability}
The specific Python codes used for training $\pi^2$ networks are available at https://github.com/NeuRonICS-Lab/Processing-in-Interconnect-Codes. 
Omnet++ codes will be available on request.

\section*{Data availability}
All datasets used for benchmarking the $\pi^2$-NN network are publicly available. 

\section*{Competing interests}
The authors declare no competing interests.

\bibliography{sn-bibliography}

\newpage
\setcounter{page}{1}

\section*{Supplementary Information for When Routers, Switches and Interconnects Compute: A Processing-in-Interconnect Paradigm for Scalable Neuromorphic AI}
\beginsupplement
\setcounter{section}{0}
\section{A simple $\pi^2$ inspired point-to-point (P2P) processor architecture}\label{secA1}
\begin{figure*}[!htbp]
\centering
\includegraphics[width=\textwidth]{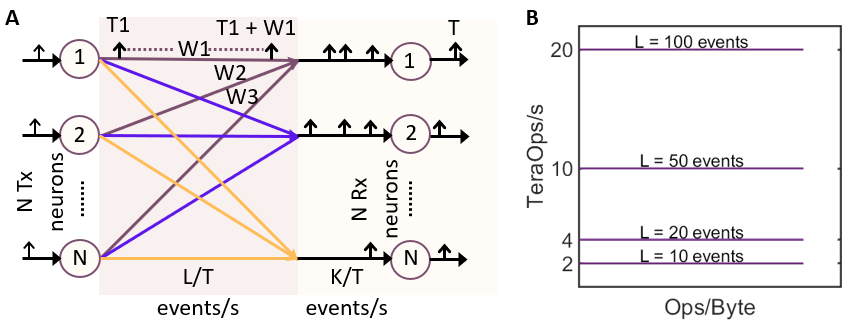}
\caption{A) $\pi^2$ inspired P2P architecture. B) Roofline model of the proposed architecture }
\label{figa1}
\end{figure*}
We propose a point-to-point $\pi^2$ network architecture (Fig.\ref{figa1}A) where every pair of neurons is connected through a path with a specific routing delay. The synaptic weights in a conventional neural network correspond to interconnect delays here. The interconnect length—or the specific routing path taken by the data—is determined by the trained synaptic weight magnitudes and is pre-defined. Thus, whenever an input neuron sends out data, it gets delayed automatically before it reaches its destination. Also, the shared interconnect bandwidth to the Rx neuron can only accommodate K events. Since the events arrive in a time-sorted manner, the first K events will inherently be accommodated at the Rx node, and the remaining events will be dropped. This approach effectively integrates processing and memory within the interconnect itself, as it inherently delays, sorts, and drops events, eliminating the need for explicit synaptic weight storage. We assume the neurons possess sufficient local memory and compute to accumulate membrane potential values and generate an event at T (output spike time).
\newline
\newline
\textbf{Performance Computation:} Consider we have an (n x n) P2P network architecture connecting n Rx and n Tx neurons. Let the computational latency of the architecture be T seconds (s) (average time taken for an output neuron to spike). The interconnect between Tx and Rx neurons has a bandwidth (BW) of 1 event/T(s).
The performance of the processor can be computed as follows:
\begin{equation}
    Ops/s = \frac{n^2  + n + n(n - K)  }T \label{eq1}
\end{equation}The number of delaying operations equals $n^2$, event dropping operations equal $(n - k)$ at every Rx node (assuming the Rx neurons only fire once in T(s)), and neuron operations equals n (at the Rx nodes—data processing and event generation). 
\newline
If the bandwidth of the interconnect is increased L times, then the performance of the processor is:
\begin{equation}
   Ops/s = \frac{Ln^2  + n + n(Ln - K) }T \label{roof}
\end{equation}
As can be observed, only increasing the number of interconnects in this architecture, or the bandwidth, can delay, drop, and transmit more events, thus improving the peak throughput of this processor architecture. Additionally, there is no access to any external memory, so the roofline model remains parallel to the X-axis, as shown in Fig. \ref{figa1}B. The latency of computation (T) will likely be further reduced, as the Rx neurons are likely to generate events faster (as they receive more data), which is bound to improve the throughput even further.
\newline
\newline
\textbf{Power utilization:}
The total energy consumption of the (n x n) P2P architecture has three main components: 
\begin{itemize}
    \item Average transmission energy ($E_{t} \ J/event$) of the events transmitted on the interconnects
    \item  Computational energy at the Rx nodes ($E_c \ J$)
    \item  Assume the input nodes are sending data at a rate of $\frac{F}{T}$ events/s. Events are lost at the input channel whenever the bandwidth of the channel cannot accommodate $F$ events in T seconds. So, the energy related to an event loss ($E_p \ J/event$) also has to be considered. 
\end{itemize}
If the interconnect between the Tx and Rx nodes has a bandwidth BW of L events/T(s), then the power consumed (W) in T seconds is:
\begin{equation}
    P_{useful} = \frac{Ln^2E_t + nE_c}{T}  \label{eq3}
\end{equation}
\newline
Considering the energy lost due to event drop $E_{lost}$ at the Tx nodes, the power lost by the system can be computed as: 
\begin{equation}
    P_{lost} = \frac{E_pn|F - L|_+}{T} \label{eq4}
\end{equation}
where $[.]_+ = \max(0,.)$ is a ReLU operation. When interconnects contribute to both computation and transmission, the transmission energy becomes an integral part of the system’s operational cost rather than an overhead. The energy/power utilization of the system can be computed to be:
 \begin{equation}
    \eta = 1 - P_{lost}/(P_{useful} + P_{lost}) \label{ee}
\end{equation}
By increasing the bandwidth of the interconnect to accommodate F events in T(s), the system achieves maximum power/energy utilization of 1.
%explaining plots in fig2
To demonstrate the implications of the $\pi^2$ model on the roofline plot (\ref{figa1}B), we arbitrarily set the variables n, T, and K to 1000 nodes, 10 $\mu$s, and 1 event. We set the values of L to 10,20,50, and 100 events and plot the roofline model based on Eq. \ref{roof}.

\section{$\pi^2$-NN Networks}\label{secA2}
How to design a neural network that uses interconnects for computing? 
\newline
To leverage interconnects for computing, we study $\pi^2$-NN, a type of neural network, consisting of neurons that are arranged in N hidden layers. The events of neurons in layer $n-1$ are sent to neurons in layer $n$. The layers are either fully connected (i.e., each neuron receives input from all neurons in the previous layer) or convolutional (i.e., connections are limited to be local and share weights).  In the following equations, an upper index refers to the layer number. The $D$-dimensional real-valued inputs $X$, such as pixel intensities, are encoded into differential event spike times as $ T^0_+ = |A+X|_+, T^0_-=|A-X|_+$ and fed as an input to $\pi^2-NN$. The synaptic weights (W) are encoded as differential synaptic delays represented as $W_+ = |B + W|_+,W_- = |B-W|_+$ for all layers. Here, A and B are arbitrary constants, and $|.| = max(0,.)$ ensures that the information represented in time is non-negative. Given the differential spike times and synaptic delays of the neurons in the previous layer $(n-1)$, the differential input spike times for the next layer $(n)$ are computed using the $\pi^2_{K}$ neuron model dynamics (Eq. \ref{k}).
\begin{equation}
    T^{n}_+,T^{n}_- = \pi^2_{K}(T^{n-1}_+,T^{n-1}_-, W^n_+,W^n_-)
    \label{k}
\end{equation}
\subsection{Algorithmic Realization of the $\pi^2_K$ Neuron in Software}\label{seca21}
The differential formulation realization of the $\pi^2_{K}$ neuron model, inspired from \cite{temp}, is as follows:  
Let \( T_{+}^{n-1},\ T_{-}^{n-1},\ W_{+}^{n},\ W_{-}^{n} \) be the differential input spike times and synaptic delays to the $\pi^2_{K}$ neuron at layer \( n \).

\begin{itemize}
    \item First, we compute the sum of the differential inputs and weights and sort them as follows:
    \begin{align}
        H1^{n} &= \text{sort}[T_{+}^{n-1} + W_{+}^{n},\; T_{-}^{n-1} + W_{-}^{n}] \quad &
        H2^{n} = \text{sort}[T_{+}^{n-1} + W_{-}^{n},\; T_{-}^{n-1} + W_{+}^{n}]
        \label{k1}
    \end{align}

    \item Then, the average of the first \( K \) elements of \( H1^{n} \) and \( H2^{n} \), denoted as
    \[
    H_+^{n} = \alpha(H1^{n}[1{:}K])/K, \quad H_-^{n} = \alpha(H2^{n}[1{:}K])/K,
    \]
    is used to compute the differential output spike times:
    \begin{align}
        T_{+}^{n} &= M + max(0,(H_-^{n} - H_+^{n})) \quad &
        T_{-}^{n} = M - max(0,(H_-^{n} - H_+^{n}))
        \label{k2}
    \end{align}
\end{itemize}

 K decides the number of inputs that contribute, and $K\leq D^{n-1}$, if the output dimension of the $n-1$ layer is $D^{n-1}$-dimensional. M is a positive constant chosen to maintain the causality between the input and output spike times (i.e., $T_+^{n} \ \& \ T_-^{n}$ are always greater than the first $K$ inputs that contributed to them). $\alpha$ is a positive scaling constant. These differential outputs are sent as inputs to the next layer. The function can be summarized by Eq. \ref{k}.
The fundamental operations of the $\pi^2$-NN type of neural networks are delaying (each input event is delayed based on its corresponding synaptic weight), time-based sorting (only the first $K$ earliest arriving events are selected for computation), and event dropping ( The remaining $(D - K)$ later events are discarded). The neurons in the $\pi^2$ network accumulate and compute the mean spike time of the first K events that arrive at it. This establishes a neural network design that leverages the inherent properties of interconnects to perform computation.
 % The learning rule for the $\pi^2_{K}$ neuron model is simple. In the context of backpropagation, the partial gradient of the output spike time T with respect to the $t_{j} $ or $w_{j}$ is $1/K$ if $(t_j + w_j)$ contributes to T, otherwise it is 0.  The learning algorithm for the $\pi^2$-NN is explained in detail in the Methods section.
 \subsection{$\pi^2$-NN can learn}\label{seca22}
\begin{figure*}[!htbp]
\centering
\includegraphics[width=\textwidth]{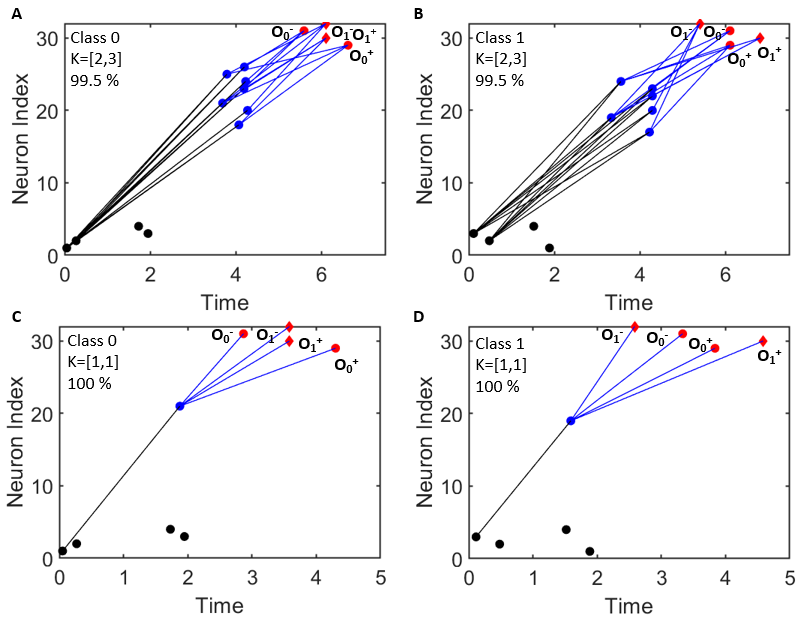}
\caption{A) The spiking activity of a 2 x 10 x 2 network trained on the XOR dataset was evaluated across input classes and values of K. The coding scheme is spatiotemporal: information is carried by the subset of neurons that activate in each layer and, within that subset, by their temporal order of spiking. Because synaptic delays allow a single neuron to be delayed by multiple offsets, a given unit can participate in multiple contributing groups that drive the next layer. Selectivity, therefore, arises from spatial sparsity (which units participate) and temporal ranking (when they arrive). This joint code is inherently combinatorial, as the number of distinct patterns grows with both the set of active neurons and their permutations in time.  A,B represent the spiking pattern for classes 0 and 1 when the K is set to be 2 and 3 at the hidden and output layer, respectively. C,D represent the spiking activity obtained when K is 1 in both layers. All the neurons are represented in a differential manner to depict their differential spiking time instants as presented in Eq. \ref{k2}. The 2 input neurons $I_0,I_1$ are represented differentially as {$I_0^+,I_1^+,I_0^-,I_1^-$}. Similarly, the hidden nodes $H_0^+,H_1^+,...,H_9^+,H_1^-,H_2^-,...,H_{9}^-$ are represented using indices 5-24, and the two output nodes for classes 0 and 1 {$O_0^+,O_1^+,O_0^-,O_1^-$} are indexed from 25-29. With K=[2,3] the network achieves a test accuracy of 99.5\% and with K set as [1,1] the network achieves a test accuracy of 100\%.  }
\label{xor}
\end{figure*}
\begin{figure*}[!htbp]
\centering
\includegraphics[width=\textwidth]{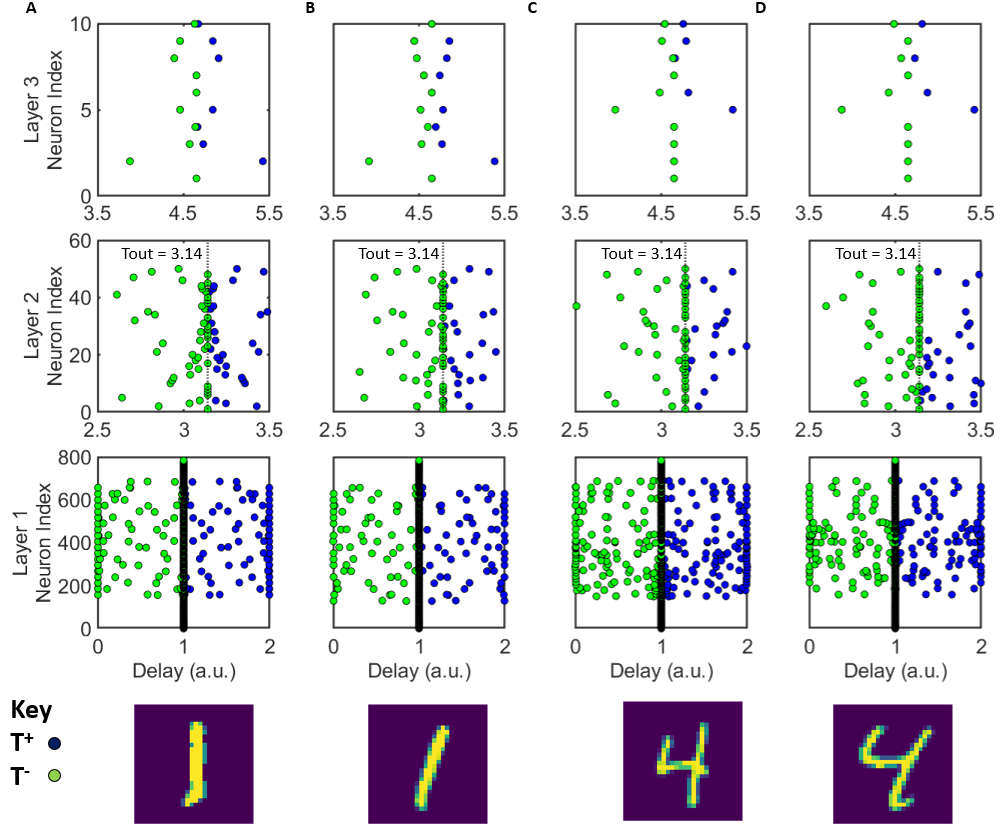}
\caption{The spatio-temporal patterns emerging from the layers of the trained $\pi^2$ network are illustrated as a raster plot. The blue and green dots represent the differential event generation times of the nodes in a layer. The activity across a population of neurons (across layers) over time for the classes 1 (A,B) and 4 (C,D) iv) represents the original image from the MNIST dataset.}
\label{figmnist}
\end{figure*}
Learning in $\pi^2$-NN requires the ability to relate the differential output spike times with both the differential input spike times and synaptic delays. In the $\pi^2_{K}$ neuron model, the contributions to $H_+, H_-$ come only from the $K$ earliest events. Consequently, the gradient of the output spike times of layer $n$ with respect to the differential weights can be computed as:
\begin{align}
\frac{\partial T_{+}^{n}}{\partial W_{i+}^{n}} &= -\frac{\alpha}{K} \mathbbm{1}(T_{i+}^{n-1} + W_{i+}^{n} \in H1^{n}[1{:}K]) + \frac{\alpha}{K} \mathbbm{1}(T_{i-}^{n-1} + W_{i+}^{n} \in H2^{n}[1{:}K])  \label{k3} \\[5pt] 
\frac{\partial T_{-}^{n}}{\partial W_{i+}^{n}} &= -\frac{\alpha}{K} \mathbbm{1}(T_{i-}^{n-1} + W_{i+}^{n} \in H2^{n}[1{:}K]) + \frac{\alpha}{K} \mathbbm{1}(T_{i-}^{n-1} + W_{i-}^{n} \in H1^{n}[1{:}K]) \\[5pt]
\frac{\partial T_{+}^{n}}{\partial W_{i-}^{n}} &= -\frac{\alpha}{K} \mathbbm{1}(T_{i-}^{n-1} + W_{i-}^{n} \in H1^{n}[1{:}K]) + \frac{\alpha}{K} \mathbbm{1}(T_{i+}^{n-1} + W_{i-}^{n} \in H2^{n}[1{:}K]) \\[5pt]
\frac{\partial T_{-}^{n}}{\partial W_{i-}^{n}} &= -\frac{\alpha}{K} \mathbbm{1}(T_{i+}^{n-1} + W_{i-}^{n} \in H2^{n}[1{:}K]) + \frac{\alpha}{K} \mathbbm{1}(T_{i-}^{n-1} + W_{i-}^{n} \in H1^{n}[1{:}K])
\label{k4}
\end{align}
Here, $i$ is the index of the $i^{th}$ input spike time and its corresponding delay. Similarly, gradients can be computed with respect to input spike times:
\begin{align}
    \frac{\partial T_{+}^{n}}{\partial T_{i+}^{n-1}} &= -\frac{\alpha}{K} \mathbbm{1}(T_{i+}^{n-1} + W_{i+}^{n} \in H1^{n}[1{:}K]) + \frac{\alpha}{K} \mathbbm{1}(T_{i+}^{n-1} + W_{i-}^{n} \in H2^{n}[1{:}K]) \\[5pt]
    \frac{\partial T_{-}^{n}}{\partial T_{i-}^{n-1}} &= -\frac{\alpha}{K} \mathbbm{1}(T_{i-}^{n-1} + W_{i-}^{n} \in H1^{n}[1{:}K]) + \frac{\alpha}{K} \mathbbm{1}(T_{i-}^{n-1} + W_{i+}^{n} \in H2^{n}[1{:}K])
    \label{k5}
\end{align}

Here, $\mathbbm{1}(J)$ is the indicator function on condition J, which evaluates to 1 if the condition J is met and to 0 otherwise. K and $\alpha$ are hyperparameters for the $\pi^2_{K}$ neuron model. Using the exact spike-based derivatives provided by the $\pi^2_{K}$ neuron model, the chain rule can be applied to compute gradients of the loss with respect to the differential weight parameters across a multi-layered $\pi^2$-NN.
\newline
We can now apply the findings above to study learning in a layered network. The output of the network is defined by the identity of the label neuron that has the maximum difference between its differential output spiking times. During training, the objective is to maximize the temporal separation between the differential output of the correct class neuron than that of all other neurons in the output layer. To facilitate supervised learning, we employ the cross-entropy loss function, applied to the difference of the differential outputs of the neurons, allowing the use of backpropagation to update the delay-encoded synaptic weights throughout all layers of the $\pi^2$-NN. Gradient descent on the loss function equation can be easily performed by repeated application of the chain rule, using the exact derivatives in equations \ref{k3}-\ref{k5}. Using the derived gradients, we train a 2 x 10 x 2 - $\pi^2$ network on the XOR dataset, and represent its spiking activity in Fig.\ref{xor}. We choose a small toy problem to better understand and illustrate the spatiotemporal spiking dynamics of $\pi^2$ networks. We generate \(1{,}000\) two-dimensional samples by drawing each coordinate uniformly from \([-1,1]\) and assigning class labels according to the XOR rule on the coordinate signs (points in opposite-sign quadrants are labeled class~1; same-sign quadrants class~0). The dataset is randomly permuted and split \(80/20\) into training and test sets with balanced classes.

%explain about traffic shaping
\section{Processing using Ethernet}\label{secA3}
\subsection{{$\pi^2$-CBS and $\pi^2$-ATS Protocol}\label{secA31}}
{The required modifications to the CBS protocol (IEEE 802.1Qav) are explicitly detailed in Fig.~\ref{cbs} and Table \ref{t1}, and the corresponding adaptations to the ATS protocol (IEEE 802.1Qcr) are described in Algorithm \ref{alg:ats}.  When the $\pi^2$-CBS and $\pi^2$-ATS algorithms are used as described, we observe a 2\% accuracy drop relative to SOTA on a LeNet-5 network trained on FMNIST. Introducing the additional scheduling mechanism described in Eq. \ref{k77},\ref{k99} of the manuscript, followed by differential encoding, restores SOTA performance. The primitives required for this scheduling mechanism—ReLU, addition, and subtraction—correspond to well-established scheduling primitives already present in Ethernet switches\cite{vc}. Future work will explore how existing scheduling protocols can be leveraged to improve accuracy and how $\pi^2$ networks can be trained to achieve high accuracy without relying on this additional scheduling logic. At present, this scheduling mechanism can be implemented as external logic interfaced with the switch datapath.}

\begin{table}[h]
\centering
{\color{red}
\begin{tabular}{p{4cm} p{4.5cm} p{4.5cm}}
\hline
\textbf{Behavior} & \textbf{Conventional CBS} & \textbf{$\pi^{2}$ CBS} \\
\hline

Transmission eligibility condition 
& Credits $\ge 0$ 
& Credits $\ge M$ (threshold)  \\[6pt]

Rate of credit accumulation when events are queued
& Credits increase at a rate defined by \texttt{idleslope}; user-specified constant 
& A counter controls \texttt{idleslope}; each packet entering the queue increments it by 1 \\[6pt]

Rate of credit decrease during transmission 
& Credits decrease at a rate defined by \texttt{sendslope}; user-specified constant 
& Transmit one event and credits drop to 0. The \texttt{sendslope} parameter can be set very high to emulate this behavior \\[6pt]

Credit reset behavior 
& If the queue is empty, credits reset to 0
& If the queue is empty, credits reset to 0 \\

\hline
\end{tabular}
}
\caption {{Comparison of conventional CBS and $\pi^{2}$ CBS mechanisms.}}
\label{t1}
\end{table}

\begin{figure*}[!htbp]
\centering
\includegraphics[width=\textwidth]{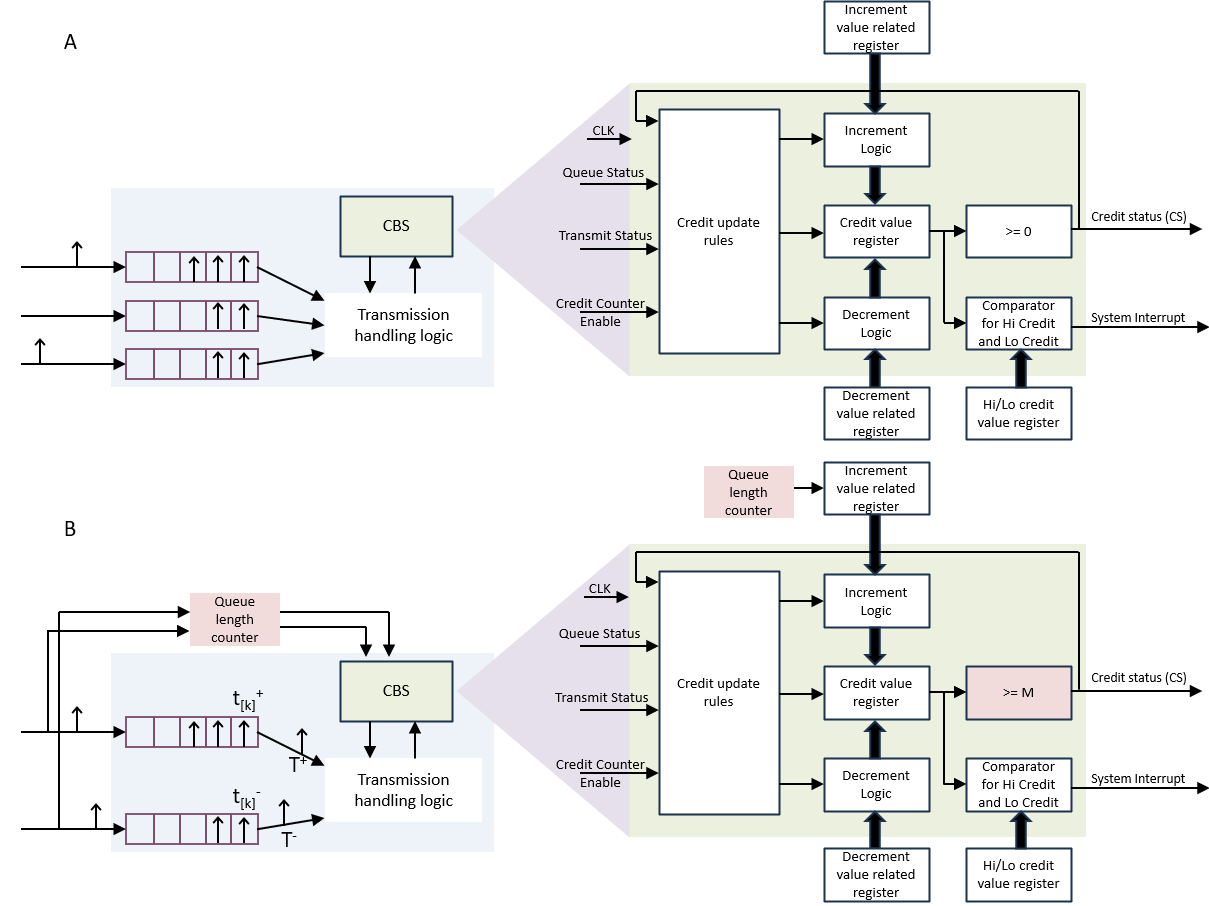}
\caption{{A) Conventional CBS implementation architecture extracted from \cite{cbs_google}. B) The $\pi^2$-CBS-based implementation—The modifications from the conventional implementation are highlighted in red. The implementation of a differential $\pi^2$ neuron (Eq. \ref{k44} of the manuscript) is presented here. As depicted, $\pi^2$ computation requires modest microarchitectural extensions to the CBS implementation—specifically programmable/non-zero credit thresholds (an extension discussed in \cite{cbs_threshold}) and queue length-dependent credit accumulation—which extends the internal credit-update logic to support $\pi^2$ computation. These changes preserve IEEE standards and Ethernet frame formats.}}
\label{cbs}
\end{figure*}
\begin{algorithm}[!htbp]
\begin{algorithmic}[1]
\color{red}{
\State /* \textbf{Parameters set during configuration} */
\State $\textbf{rate} \rightarrow \infty$
\State $\textbf{burstSize = 0}$
\State $\textbf{MaxTime = w}$
\State /* Initialization */
\State $T_{\text{eligibility}} = 0$
\State $T_{\text{bucketFull}} = 0$
\State $T_{\text{groupEligibility}} = 0$
\State $T_{\text{bucketEmpty}} = -(burstSize/rate)$

\State /* Frame Processing */
\State $D_{\text{lengthRecover}} = frame.length/rate$
\State $D_{\text{emptyToFull}} = burstSize/rate$
\State $T_{\text{shaperEligibility}} = T_{\text{bucketEmpty}} + D_{\text{lengthRecover}}$
\State $T_{\text{bucketFull}} = T_{\text{bucketEmpty}} + D_{\text{emptyToFull}}$
\State $T_{\text{eligibility}} = \max(T_{\text{arrival}},$
\State \hspace{2em} $T_{\text{groupEligibility}},$
\State \hspace{2em} $T_{\text{shaperEligibility}})$

\State /* Shaping */
\If{$T_{\text{eligibility}} \le (T_{\text{arrival}} + MaxTime)$}
    \State $T_{\text{groupEligibility}} = T_{\text{shaperEligibility}}$
    \State $T_{\text{bucketEmpty}} =
    (T_{\text{eligibility}} < T_{\text{bucketFull}}) ?$
    \State \hspace{2em} $T_{\text{shaperEligibility}} :$
    \State \hspace{2em} $T_{\text{shaperEligibility}} + T_{\text{eligibility}} - T_{\text{bucketFull}};$
    \State \textbf{$AssignAndProcess(frame, T_{\text{arrival}}  + MaxTime)$} \textbf{(modified)}
\Else
    \State $Discard(frame);$
\EndIf
}
\caption{{: In $\pi^2$-ATS, the standard ATS eligibility computation and residence-time admission check are preserved without modification. By configuring the committed rate to a large value and setting the burst size to zero, eligibility reduces to the arrival timestamp. After successful admission, frames are scheduled at the precomputed value (Tarrival+MaxTime), reinterpreting the existing ATS residence-time bound as a synaptic delay. As a result, there is no additional energy consumption beyond that of the conventional ATS datapath. No frame formats or scheduling rules defined by IEEE TSN are altered. While conventional ATS implementations transmit frames immediately upon eligibility, $\pi^2$-ATS deliberately schedules transmission at the residence-time bound (Tarrival+MaxTime), which remains fully compliant with IEEE TSN timing constraints. The conventional ATS protocol is adapted from \cite{ats1,ats2}, and the modifications required for $\pi^2$ processing (Eq.\ref{k441},\ref{k442}) are highlighted with the keyword modified. }}
\end{algorithmic}
\label{alg:ats}
\end{algorithm}
\subsection{Vision for large-scale systems}\label{secA32}
\begin{figure*}[!htbp]
\centering
\includegraphics[width=\textwidth]{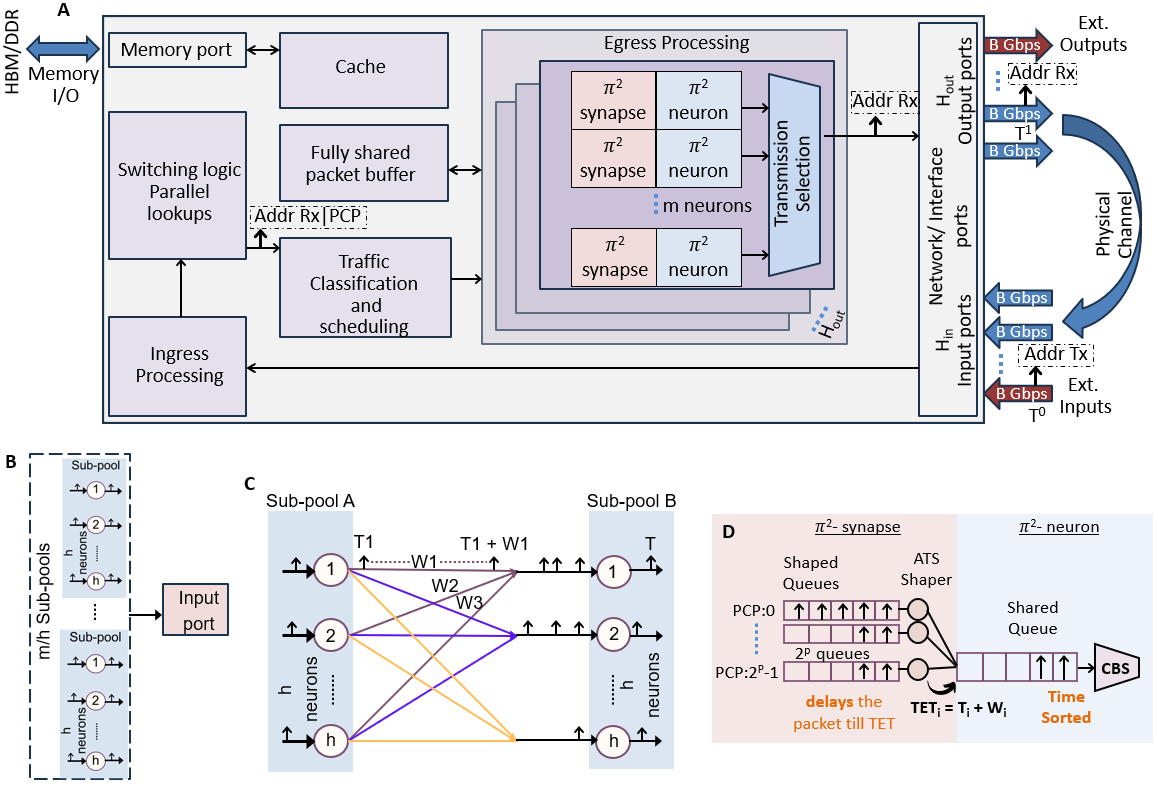}
\caption{A) A subpool of h neurons is fully connected to another subpool of h neurons.  B) A port is shared by m neurons. C) The proposed $\pi^2$-NN implementation using an Ethernet switch.  The Ethernet switch can support $n=mH_{out}$ neurons in total.   }
\label{fig10}
\end{figure*}
We strategize and analyze an implementation for a processing system consisting of n neurons, communicating through an Ethernet switch (with H ports each supporting a bandwidth B), as shown in Fig. \ref{fig10}A. $m$ neurons share a port. There are sub-pools (among these m neurons), each consisting of h neurons. Thus, a port is shared by m/h sub-pools (Fig. \ref{fig10}B).  We propose an architecture where a sub-pool of h neurons is fully connected to another sub-pool of h neurons (Fig. \ref{fig10}C). Each sub-pool-pool connection consists of $h^2$ synapses. For each of these synaptic connections, p bits representing the synaptic delays have to be stored. The number of sub-pools in the core is $n/h$. The neuron's address consists of the port address it corresponds to ($\log_2(n/m)\ \text{ bits}$), the pool it belongs to ($\log_2(m/h) \ \text{ bits}$), distinct neuron address in that pool ($\log_2(h) \ \text{bits}$) - which amounts to a total of $\log_2(n) \ \text{ bits}$.
Features of this system include: 
\begin{itemize}
      \item The network activity is controlled by the Ethernet switch. It is configured to have $H_{in}$ and $H_{out}$ input and output ports, respectively. Each input and output port of the switch is associated with a channel that is shared among m neurons. Consequently, the total number of neurons in the system is given by $n = mH_{in}$ neurons. Here, we assume that $H_{in}=H_{out}$ to balance the input and output traffic.
    \item The external sensory input (purely Tx nodes) comprising of $m$ input neurons communicates data through one port of the switch. Similarly, there is a dedicated port supporting $m$ output neurons (purely Rx nodes). The remaining intermediate neurons (Tx and Rx nodes) communicate through the $H-2$ ports of the switch. All the neurons have an average spiking frequency of $F_{spk}$ events/s.
    \item The processing starts when the m Tx nodes of the external sensory input start sending event-based data to the switch.   Each Tx event communicated to the switch comprises the source address of the Tx neuron that generated it, represented 
    by $log_2(n)$ bits.
     \item The switch receives events through its input port and processes them. A routing/MAC table containing destination addresses (fanout of the Tx event) and the weights of synaptic connections is stored either within the switch or in external memory. Each neuron supports h fan-in/fan-out connections.
     \item The Tx address is decoded, the routing table is accessed, and events with appropriate destination addresses (Rx address) are generated. Similar to the previous implementation, a PCP tag (p bits) is added to the events representing the synaptic delay value. The events are directed to their corresponding destination ports by the traffic classification and scheduling logic.
      \item Every output port is associated with $m$ set of shaped queues (Fig. \ref{fig10}D), one for each neuron that shares that port. A set of shaped queues consists of $2^p$ queues, each one for each distinct PCP value. Events associated with the same PCP tags for a neuron are dropped in the same queue. So, the events will inherently be arranged in the ascending order of their arrival times. Thus, the queue selection is based on the Rx neuron address and the PCP tag. 
    \item The ATS shaper controls each shaped queue. Based on the synaptic delay (PCP tag), a TET tag is assigned to the events to schedule their transmission to the shared queue of the neuron by the ATS shaper. When the appropriate TET value of an event is reached, it is released to the shared queue of that Rx neuron.
     \item A modified CBS shaper controls the shared queue of the neuron. As events arrive, the credits are accumulated. When the CBS-controlled gate opens for transmission, an event encoded with the Rx neuron address, consisting of $log_2(n)$ bits, is forwarded to the output ports of the switch. The remaining events are dropped (TTFS encoding).  This event now becomes the next Tx event processed by the switch. 
     \item This process continues until the m output neurons (Rx nodes) receive data, after which there will be no communication between the neurons. 
    \item For the power analysis, we assume the weights are uniformly distributed  (a similar assumption was taken in \cite{truenorth}). 
\end{itemize}

\textbf{Bandwidth requirements at the input/output port:} On average, the number of neurons (m) that can share a port or channel that can support a bandwidth of B bits/s can be computed by solving the following inequality.
\begin{equation}
    B \geq F_{spk}mlog_2(n)
    \label{bin}
\end{equation}
\textbf{Memory requirements:} Say we have a look-up table in the switch storing the connectivity information of every pool-pool pair and their corresponding synaptic delays. Here, we assume that every input port is associated with a memory block storing the connectivity information of the m neurons that share that port. Once a Tx event arrives at an input port, its pool number is extracted ($log_2(m/h) \ bits$), and the LUT is accessed to retrieve the destination port ($log_2(n/m) \ bits$) and the destination pool address ($log_2(m/h) \ bits$) it is associated with. The Tx event is fully connected to every neuron in the destination pool.  For every pool-pool connection, we have to store $h^2p$ bits. Thus, for one pool-pool connection, the LUT stores $h^2p + log_2(nm/h^2) \ bits$. Every port is associated with m/h pools, and there are $H_{in}$ ports. Thus, the total memory required to store the LUT for all the input ports will be: 
\begin{equation}
    M \geq nhp + (n/h)log_2(mn/h^2) \ bits
    \label{mem}
\end{equation}
\textbf{Shared queue size at every neuron:} Every neuron is associated with a shared queue with a capacity to support K events. The total memory required to accommodate the shared queues is $nKlog_2(n)$ bits. If we consider a shared buffer system (helps in reducing the memory requirement), where the total incoming traffic to the n shared queues in the switch is $nKlog_2(n)F_{spk}$ bits/s, and the maximum time the CBS gate is closed for is $T_{out}$ time units, the memory required for the shared queues will be 
\begin{equation}
    Q_{nsize} = nKlog_2(n)F_{spk}T_{out} \ bits(time\ units/s)
    \label{qneuron}
\end{equation}
\textbf{Shaped queue size at the output ports:} Every output port has $2^pm$ shaped queues, one for each distinct PCP code (p bits), and neuron it supports (m neurons share a port, and each neuron has a set of $2^p$ queues). The traffic associated with an Rx node is a function of its fanin and rate of arrival of input events ($hF_{spk} \ events/s$). However, since only the first K events matter,  each of the queues associated with a neuron has to accommodate an input traffic of $KF_{spk} \ event/s$. The average time an event spends in a queue is a function of the synaptic delay value. If the $2^p$ distinct PCP tags correspond to a delay of $0-(2^p-1)$ time units, then the average time an event spends in the queue is (assuming uniform distribution):
\begin{equation}
    T_{avg} = 2^p(2^p-1)/2 \ time \ units
\end{equation}
The required queue size will be:
\begin{equation}
    QN_{size} = T_{avg}KF_{spk}(log_2(n)+n_t) \ bits(time\ units/s)   
\end{equation}
% Since every port is connected to m distinct neurons, only $log2(m)$ bits have to be stored (port address in common). The extra 32 bits are to store the TET value of the events. 
Here, $n_t$ is the number of bits required to store the TET (transmission eligibility time). If more than $KF_{spk} \  events/s$ occur, then they will be automatically dropped. If the synaptic delay values are uniformly distributed among the $2^p$ queues of a neuron, then the queue size can be further reduced by $2^p$ times. Under this assumption, the total queue memory required in bytes across all the output ports is:
\begin{equation}
    Q_{size} = T_{avg}nKF_{spk}(log_2(n) + n_t)/(8*2^p) \ B(time\ units/s)   \label{qsize}
\end{equation}
% \textbf{Bandwidth requirements at the output port channel:} On average, the traffic forwarded to the output port of the switch should satisfy the following constraint: 
% \begin{equation}
%     B \geq 2^pKF_{spk}mlog_2(n)
% \end{equation}
% Here, a factor of accounts for the $2^p$ queues allocated to each of the m neurons that share a port. The input traffic at every queue is $KF_{spk} \ events/s$, and we need $log_2(n)$ bits to address a neuron. If the synaptic weights are uniformly distributed then, 
% \begin{equation}
%     B \geq KF_{spk}mlog_2(n) \ bits/s
%     \label{bout}
% \end{equation}
\textbf{Throughput:} Neuromorphic systems are usually compared in terms of synaptic operations per second (SOPS). 
The SOPS for this system is related to its network activity and can be computed as a product of the average firing rate of the neurons and the average number of active synapses \cite{truenorth2,spinnaker}. 
% \begin{equation}
%     T_{sops} = SF_{spk}hn \ SOPS
%     \label{sops}
% \end{equation}
% The energy per synaptic event is the commonly reported metric \cite{truenorth}:
% \begin{equation}
%     \eta = \frac{P_{total}}{T_{sops}} \ J/event
% \end{equation}
 The throughput of the $\pi^2$ system can be computed based on its three fundamental operations: number of effective delay operations performed, number of packets transmitted by the interconnect system (neuron activity), and number of packet drops at the queues of the output ports (at the neuron level).  
\newline
The number of effective synaptic (delay) operations supported by the interconnect system can be computed as (assuming uniform synaptic delay distribution)
\begin{equation}
    T_s = nKF_{spk} \ events/s
\end{equation}
The number of events transmitted by the interconnect systems (neuron activity) is 
\begin{equation}
    T_n = nF_{spk} \ events/s
\end{equation}
The number of event drops at the shaped and shared queues of the output ports of the switch is (where $h$ is the average number of received events, and K is the number of events used for output computation).
\begin{equation}
    T_d = nF_{spk}(h-K) \ events/s
\end{equation}
The peak throughput of the system can be computed as:
\begin{equation}
    T_{peak} = nKF_{spk} + nF_{spk} + n(h-K)F_{spk} \ Operations/s
    \label{tpeak}
\end{equation}
It is important to note that these operations are primarily implemented using the interconnect system (Ethernet switches, queues).
\newline
\newline
\textbf{Power analysis}:  The switches, interconnect transmission power, and memory access dominate the system's network power consumption. 
If the interconnect modules to the switch consume $a \ J/bit$ then the interconnect power can be calculated as:
\begin{equation}
    P_i = aBH \ Watts
\end{equation}
If it takes $b \ J/bit$ to access the memory related to the routing table and queues at the output port, then the power consumption to access $V \ bits/s $ is
\begin{equation}
    P_m = bV \ Watts
\end{equation}
 The total power of the system, consisting of a switch, memory, and interconnects (excluding the neurons), can be computed as: 
\begin{equation}
    P_{total} = P_s + P_i + P_m \ Watts
    \label{ptotal}
\end{equation}
Here, $P_s$ is the power consumption of a switch. Large-scale neural simulations are communication-bound, rather than compute-bound \cite{spinnaker2}. Here, we assume that the transmission power dominates the computing power of the neurons \cite{stanojevic2024}, specifically for networks at scale. 
\newline
\newline
\textbf{Current trends in the evolution of Ethernet switches:}
51.2-Tbps Co-Packaged Optics Ethernet Switch Platform is one of the recently launched products by multiple companies such as CISCO, Broadcom, etc \cite{cisco}.
CISCO's G200 switch consists of  512 x 100GE Ethernet ports on one device. The Marvell Teralynx 51.2T Ethernet Switch consumes 1 watt per 100 gigabits-per-second of bandwidth \cite{marvel}. Similarly, the Broadcom's Tomahawk 4 Switch consumed a total power of 450 W, producing 1.76 W per 100 gigabits-per-second of bandwidth \cite{25.6}. Considering 1W/100Gbps, the power consumption of a 51.2 Tbps switch ($P_s$) will be 512 Watts. 
 In recent technology, Co-packaged optics (CPO) has evolved as a solution to meet the growing demand for data movement. CPO with the Ethernet switch ASICs presents distinct benefits in terms of bandwidth, size, weight, and power consumption \cite{cpoprogress,cpo}. The power consumption of pluggable optical
modules varies from 15 pJ/bit to 20 pJ/bit. However, the power consumption of CPO systems can be decreased to a range of 5 pJ/bit to 10 pJ/bit \cite{cpoprogress}. Recently, Broadcom presented the performance data of  TH5-Bailly CPO 51.2-Tbps Ethernet switch delivering 5W/800 Gbps interconnect power \cite{broadcommcpo}, which amounts to 6.25 pJ/bit. 
Assuming a CPO-based interconnect power density of 6.25 pJ/bit, the total interconnect power consumption is computed as:
\begin{equation}
    Pi = 6.25 pJ/bit * 51.2 \ Tbps = 320 W
\end{equation}
The total power consumption of the network infrastructure ( switch + interconnect) from Eq. \ref{ptotal} is 832W.
\newline
\newline
\textbf{Neuron in a core:}
Assuming a spiking frequency of 10 Hz  (usual metric reported for the human brain), average fan in and fanout of 1K, for n neurons with 4-bit quantized weights \cite{quantize} (It can be inferred from the robustness analysis that 4 bits is sufficient to maintain competitive accuracy), we solve Eq. \ref{bin} to compute 
the total number of neurons in a core to be 71 B neurons with each port being shared by 277 M neurons. The system supports 71 trillion synapses.  We focus on the simplistic case of K=1 and K = 10 networks for ease of implementation and $H_{in} = H_{out}$, such that the average incoming and outgoing traffic are matched.
\newline
\newline
\textbf{Memory required to store the routing table:}
The memory required to store the synaptic information scales with the number of synaptic connections the system supports. However, the current Ethernet switches support a shared on-chip memory in the range of MB only \cite{25.6}. To support 71 trillion synaptic connections, external memory access is imperative. The total memory required by this system is around 35.5 TB (using Eq. \ref{mem}). These days, industrial switches and routers incorporate high bandwidth memory (HBM) as packet buffers to store the packets in case of excess traffic \cite{broadcommHBM,ciscohbm}. High
Bandwidth Memory (HBM) is a game-changing
technology that allows high-bandwidth switches to be
equipped with deep buffering capacity \cite{ciscohbm}. The state-of-the-art Ethernet switches are equipped with HBM, which serves as packet buffers in case of excess traffic \cite{ciscohbm,broadcommHBM}, though they are in the range of Gbytes \cite{ciscohbm}. Assuming the Ethernet switch is equipped with a larger HBM capacity to store and communicate the synaptic information, we compute the power cost incurred in HBM memory access and signaling. The projected energy for DRAM access and interface energy was 4.5 pJ/bit in 16nm technology \cite{hbm_energy2}. HBM has relatively good energy efficiency (about 3-5pJ/b) compared to other DRAMs \cite{hmb_energy3,hbm_energy2,hbm_energy4,hbm_energy5,hbm_energy6}. 
For every event, the destination port, destination pool address, and the delays of the h synaptic connections have to be accessed. The number of bits to be accessed is:
\begin{equation}
    m_{access} = hp + log_2(m/h) + log_2(n/m) \ bits 
\end{equation}
For $nF_{spk}$ events/s, the memory interconnect bandwidth has to support $nF_{spk}m_{access}$ bits/s. We assume that, for all the Fspk events/s, the memory is accessed only once. If it takes 3pJ/bit for memory access and signaling, then 858 W is consumed by the routing table memory. 
\newline
\newline
\textbf{Queue capacity:} To support large-scale simulation, the total shared memory required to support the shaped queues at the output ports for $K=1$ implementation is 45.3 MB (using Eq. \ref{qsize}) if we consider the synaptic delay to be in the range of microseconds and $n_t$ to be of 32 bits. (The current latency of the Ethernet switches is in the range of 100s of nanoseconds \cite{25.6}.)
If we consider $T_{out}$ in the range of 50 microseconds, then the memory required for the shared queues is 160 MB (Eq. \ref{qneuron}).
% The number of times the total output port queues memory will be accessed for storing $log2(m) \ bits$ will be $KF_{spk}8n$ accesses/s. 
The state-of-the-art Ethernet switches already support sufficient memory capacity for packet buffers \cite{ciscohbm,25.6} in the range of 100MBs. For K=10 networks, 10x the memory capacity will be required, in which case, HBMs have to be used (current switches support GBytes of HBM storage). 
% With the support of HBM-based packet buffers \cite{ciscohbm}, and using 3 pJ/bit energy efficiency, the power consumed in this process will amount to 478 Watts. 
\begin{table}[h]
\caption{Trends in technological advancements for interconnect systems}\label{tab1}
\begin{tabular*}{\textwidth}{@{\extracolsep\fill}lcccccc}
\toprule%
Year & SE(pJ/bit) & IE(pJ/bit) & HBME(pJ/bit) & SP (Tbps)  \\
\midrule
2020  & 17.58 & 8.8  & 5 & 25.6 \\
2022  & 10    & 6.25 & 3 & 51.2  \\
2034  & 1     & 1    & 1 & 4096 \\
\botrule
\label{table2}
\end{tabular*}
\footnotetext{SE - energy efficiency of switch, IE  - energy efficiency of interconnect, HBME - energy efficiency of HBM, SP - switch throughput in Tbps  }
% \footnotetext[1]{Sparsity among n neurons (S)}
% \footnotetext[2]{Example for a second table footnote.}
\end{table}
\begin{table}[h]
\caption{Performance projection numbers for $\pi^2$ network implementation}\label{tab1}
\begin{tabular*}{\textwidth}{@{\extracolsep\fill}lcccccc}
\toprule%
Year & Synapses(T) & Throughput (petaOp/s) & $P_{total}$ (kW) & EE(pJ/Op)  \\
\midrule
2020  & 36   & 0.365  & 1.41     & 3.85 \\
2022  & 71   & 0.71   & 1.69     & 2.37 \\
2034  & 4859 & 48.65  & 27.78     & 0.57 \\
2034 (brain-scale) & 100  & 1 & 0.548  & 0.55 \\
\botrule
\end{tabular*}
\label{table3}
\footnotetext{EE - energy efficiency of the interconnect system assuming K=1.}
% \footnotetext[1]{Sparsity among n neurons (S)}
% \footnotetext[2]{Example for a second table footnote.}
\end{table}
\subsection{{Analyzing the effects of Ethernet header overheads and higher synaptic bit precisions}}\label{secA33}
{The above projections should be interpreted as an architectural feasibility and scalability analysis of a $\pi^{2}$ neuromorphic processor implemented on an Ethernet switch with external memory (DRAM or HBM). This analysis assumes that each event contains only the source address—$\log_2(n)$ bits (approximately 5 bytes for 71B neurons)—as is typical in neuromorphic systems. In practice, Ethernet frames introduce additional header overhead: the minimum frame size is 64 bytes, with a 6-byte source address field and an optional 4-byte VLAN tag, of which 3 bits encode the PCP field. Higher-precision synaptic delays can be embedded within the VLAN ID field, allowing delay information to be carried without increasing the overall frame size.
\newline
If we consider the header, it will have a direct impact on the neuron count and performance projections. Specifically, Eq. \ref{bin} gets modified to:
\begin{equation}
    B \geq 512F_{spk}m
    \label{bin_e}
\end{equation}
Solving Eq.\ref{bin_e} with $F_{spk}=10Hz$ and 100 Gbps bandwidth leads to $m=19.5M$ neurons sharing a port (14x reduction). Accordingly, the total neuron count reduces to $5B$ neurons, and the total memory required to store the look-up tables reduces to 2.5 TB.  Following a similar analysis as above, we report the energy, throughput and performance projections with the header constraint in Table \ref{table3_e}.  
\begin{table}[h]
\color{red}{
\caption{Performance projection numbers for $\pi^2$ network implementation with Ethernet header overhead}\label{tab1}
\begin{tabular*}{\textwidth}{@{\extracolsep\fill}lcccccc}
\toprule%
Year & Synapses(T) & Throughput (petaOp/s) & $P_{total}$ (kW) & EE(pJ/Op)  \\
\midrule
2020  & 2.5   & 0.025  & 0.675     & 29 \\
2022  & 5   & 0.05   & 0.83     & 17.8 \\
2034  & 400 & 4  & 8.2     & 2.44 \\
% 2034 (brain-scale) & 100  & 1 & 0.548  & 0.55 \\
\botrule
\end{tabular*}
\label{table3_e}
\footnotetext{EE - energy efficiency of the interconnect system assuming K=1.}
% \footnotetext[1]{Sparsity among n neurons (S)}
% \footnotetext[2]{Example for a second table footnote.}
}
\end{table}
}

\newpage
\section{Simulator implementation of  $\pi^2$-NN using traffic shaping protocols}\label{secA4}
% \begin{figure*}[h]
% \centering
% \includegraphics[width=0.8\textwidth]{Figures/TEMP_TSN_1.PNG}
% \caption{a) Implementation Strategy 3: Representation of the TEMP network implemented using the distributed TSN computing paradigm. This implementation exploits the existing functionalities of ethernet routers for TEMP computation.  The transmitter and receiver nodes are TSN devices that are compatible with the TSN protocols. Each TSN node receives the sensor inputs and generates events (frames) with the corresponding PCP tags (synaptic weights are stored in memory) and destination addresses at specific time instants. The input spikes from the Tx nodes are first routed to the TSN switch implementing the asynchronous time shaper (ATS) algorithm, and the TEMP function is implemented by another TSN switch which follows credit-based traffic shaping (CBS).    }
% \label{fig:fig4}
% \end{figure*}
\begin{figure*}[h]
\centering
\includegraphics[width=0.7\textwidth]{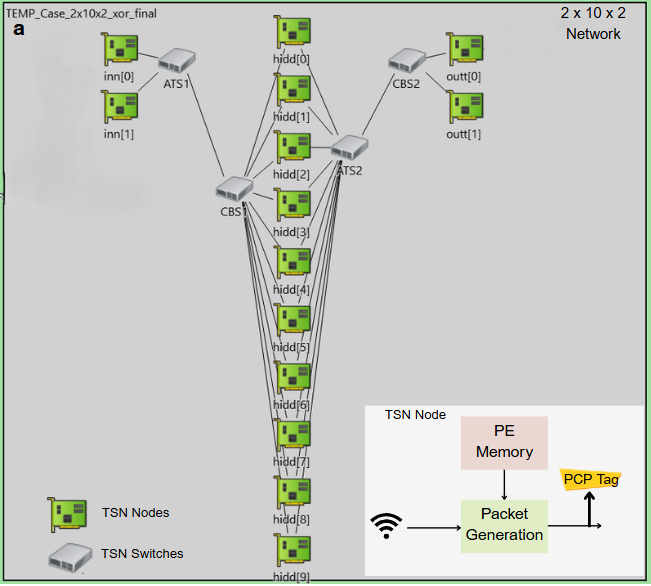}
\caption{a) $\pi^2$ network simulation on OMNET++:  This implementation exploits the existing functionalities of Ethernet routers for $\pi^2$ computation.  The transmitter and receiver nodes are TSN devices that are compatible with the TSN protocols. Each TSN node receives the sensor inputs and generates events (frames) with the corresponding PCP tags (synaptic weights are stored in memory) and destination addresses at specific time instants. The input spikes from the Tx nodes are first routed to the TSN switch implementing the asynchronous time shaper (ATS) algorithm, and the membrane potential accumulation is implemented by another TSN switch, which follows credit-based traffic shaping (CBS). An instantiation of a 2x10x2 network is portrayed here for illustration purposes. Similarly, we run a 784x50x10 network using the simulator.}
\label{fig:tsn_xor}
\end{figure*}
% \begin{figure*}
% \centering
% \includegraphics[width=0.9\textwidth]{Figures/TEMP_TSN_2.PNG}
% \caption{ a) The internal structure of the ATS-TSN switch - At the input port, transmission eligibility times (TET) of the packets are calculated by the ATS meters. Each egress port consists of 8 queues. The packets are dropped into the queues based on their PCP values (0-7) and are stored (delayed and time-sorted) there till their TET is reached. When the TET is reached, the ATS gates open, and the packets are directed to the ingress port of the CBS-based switch. b) The internal structure of the CBS-TSN switch   }
% \label{fig:fig5}
% \end{figure*}
% In this section, we describe how the existing functionalities of ethernet switches that interconnect multiple nodes in a system can be simultaneously used for computing and communicating \cite{omnet_tsn}. 
% \newline
We present a simulation model of the proposed $\pi^2$ architecture on the OMNET++ platform, which is a C++ framework that enables the construction of network simulators and has extensive libraries to support the TSN protocol \cite{omnet_tsn}. OMNET++ already has built-in code to demonstrate various traffic shaping mechanisms relevant to TSN networks. We implement a 784x50x10 $\pi^2$ network trained on the MNIST dataset \cite{temp} as a proof of concept.  
\newline
The main components of the simulation networks are as follows:
\begin{itemize}
    \item Node: Each node (Tx and Rx) consists of an input port, through which it receives event data from external sources (sensor) or other nodes in the system, a packet generation logic, and a local memory where it stores the synaptic information (destination addresses and synaptic delays). It outputs a packet with the appropriate destination address and a PCP tag (representing the delay value) attached to it. Usually, the Ethernet frames consist of a priority code point (PCP) tag, which is used to identify their traffic class \cite{nesting}. The nodes are simulated as TSN (Time-sensitive networking) device components on OMNET++ (as portrayed in Fig. \ref{fig:tsn_xor}) \cite{omnet_tsn}.  
    \item Ethernet switch: The Ethernet switches receive packets from the nodes through their ingress ports, filter, and process them with the help of traffic-shaping algorithms. The idea of traffic shaping is to classify the incoming packets depending on the QoS requirements and schedule their departure through the output ports accordingly. The output/egress ports of the switch support multiple queues (FIFOs) that buffer the packets until they are ready for transmission.
\end{itemize}
The simulation works as follows:
\begin{itemize}
    \item The input spikes from the Tx nodes are first routed to the ATS switch, implementing the asynchronous time shaper (ATS) algorithm. The main components of the ATS switch are portrayed in Fig. \ref{fig:tsn_ats}. Based on the PCP value of the frame, the classifier routes it through the appropriate gate. The ATS meter computes the TET value of the frame based on the PCP tag value.  We modify the equation to compute the TET for each event, based on the PCP tag (synaptic delay) associated with each frame. So, if the event with a PCP tag value of $W_i$ enters the switch at time $T_i$, then the TET for the frame is computed as:
 \begin{equation}
     TET = T_i + W_i
 \end{equation}
Usually, the PCP is a 3-bit code and can support 8 different traffic classes \cite{nesting}. Accordingly, each egress port of the switch consists of 8 queues. The events are dropped into the queues based on their PCP tag values, buffered until their TET is reached, and then transmitted to the ingress port of the TSN-CBS switch. Inherently, the events in the queues of the egress ports will be arranged in ascending order of their TET values. 
\item The main components of the CBS switch are portrayed in Fig. \ref{fig:tsn_cbs}. Every egress port of the CBS switch is connected to a particular Rx node. Each of these egress ports is associated with two queues (to support differential computing). The traffic shaping algorithm decides when the gate for the queue opens for event transmission. 
The CBS switch implements a modified version of the credit-based shaping algorithm (Fig. \ref{fig:tsn_cbs}). We modify the credit calculation algorithm to suit the $\pi^2_{K}$ neuron dynamics. The modified CBS algorithm computes the credits (membrane potential) at every local time instant with a slope defined by the number of events (maximum K packets are stored; the remaining are dropped due to congestion) stored in its corresponding queue. When the credit value reaches a threshold (gamma), the CBS gate of that queue opens, and a packet is directed from the queue to the corresponding Rx node. The remaining packets in the queue are dropped (non-linearity).
\end{itemize}
\begin{figure*}
\includegraphics[width=0.7\textwidth]{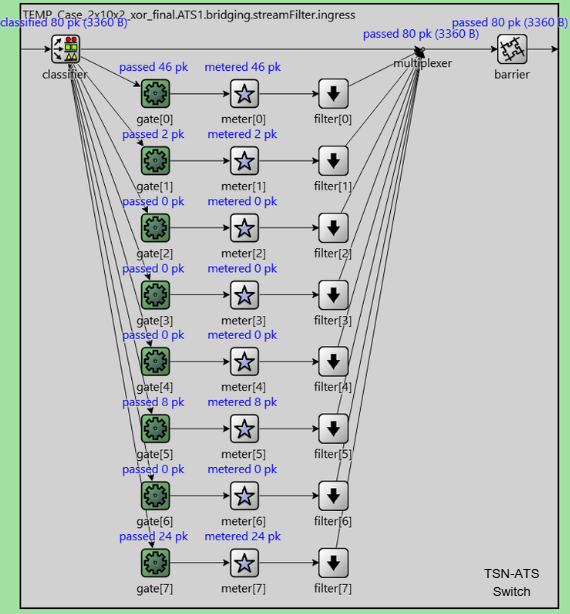}
\centering
\caption{The inner architecture of the ATS switch - Based on the PCP value (3-bit), the classifier routes the frame through the right gate. The ATS meter computes the TET value and allows frame transmission when the TET value is reached. The snapshot of this ATS mechanism at work between the input and hidden layer is portrayed here. }
\label{fig:tsn_ats}
\end{figure*}
\begin{figure*}
\includegraphics[width=1\textwidth]{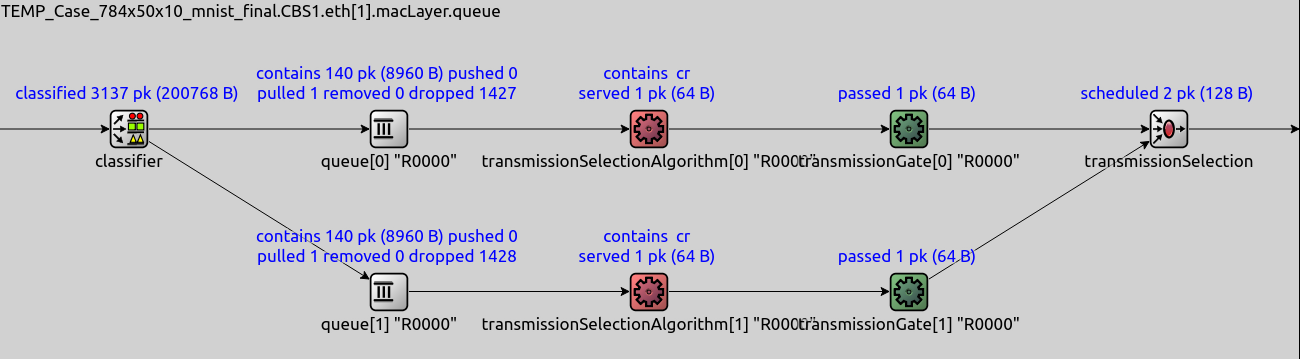}
\centering
\caption{The inner architecture of the CBS switch - The hidden layer is constrained to K=140. Thus, the shared queue (capacity) is configured to hold only 140 frames. The frames will inherently arrive in a time-sorted manner in the queues. In case of congestion, the other frames are dropped as displayed. Every hidden layer node receives 1568 (784x2) frames from the input layer. Out of which 140 are contained, one frame is transmitted, and the remaining are dropped due to congestion in the queue[0] and queue[1].}
\label{fig:tsn_cbs}
\end{figure*}
\begin{figure*}
\includegraphics[width=1\textwidth]{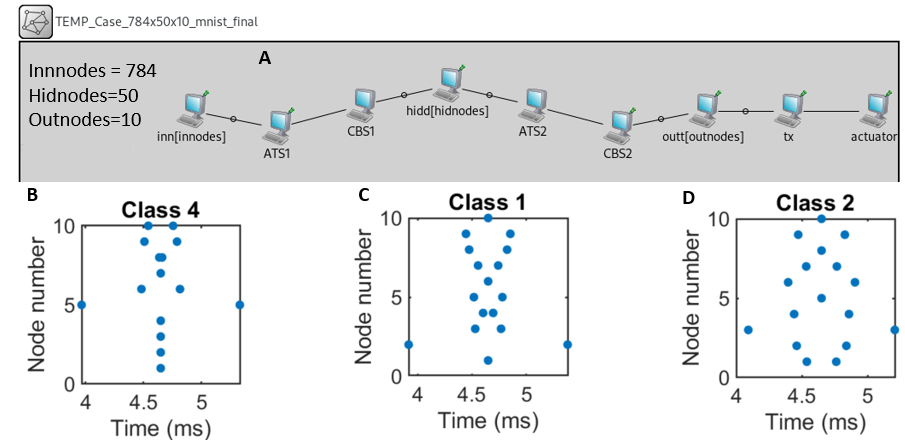}
\centering
\caption{A) A 784x50x10 network is simulated using the proposed traffic shaping protocols and Ethernet switches on the OMNETPP software. B,C,D) The spatio-temporal patterns for input of different classes are extracted from the simulator and portrayed in the figures. The index of the node that generates a packet/frame first (earliest) denotes the class the input image belongs to. The output spike times exactly match the software simulations of the quantized $\pi^2$ network. }
\label{fig:tsn_mnist}
\end{figure*}
 We trained a 3-layer $\pi^2$ based network on the MNIST dataset, quantized the weights (delays) to 3 bits (to map to 3-bit PCP value), and performed inference on the OMNET++ simulator. 
 This experiment gives an idea of an implementation strategy where the fundamental operations of the $\pi^2$ model, namely delaying, sorting (in TSN-ATS switch), event dropping, and thresholding (in TSN-CBS switch), can be mapped to existing traffic shaping protocols of Ethernet switches.  
 The spiking times of the 10 output nodes for different images are extracted and used for classification as shown in Fig. \ref{fig:tsn_mnist}.
 This implementation on the simulator provides a simplistic view to demonstrate how Ethernet switches can be used for computation and communication simultaneously using traffic-shaping protocols.
 \newline
 \newline
 \section{Effect of weight distillation}\label{secA5}
A conventional MAC-based ResNet-9 network is first trained on the CIFAR-10 dataset, and its weights are then ported to the $\pi^2$-ResNet-9 architecture. The $\pi^2$ network is subsequently fine-tuned to achieve the baseline competitive accuracy. However, when the weight distillation strategy is omitted, the $\pi^2$- network experiences an accuracy drop of approximately 1-2\%.
\begin{figure*}
\includegraphics[width=0.9\textwidth]{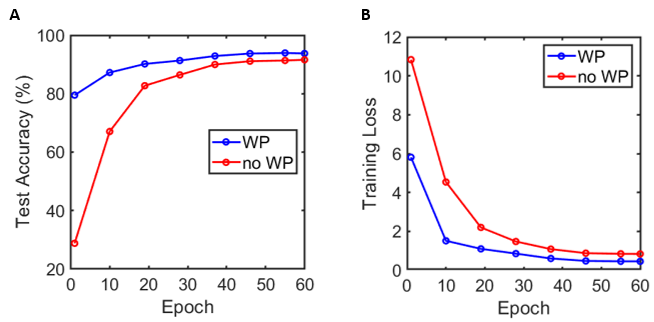}
\centering
\caption{ The impact of weight porting (distillation) (WP) on training dynamics and generalization performance of the ResNet-9 network trained on the CIFAR-10 dataset is illustrated.
(A,B) Test accuracy (\%) and training loss over training epochs with and without weight distillation from a trained conventional teacher network. }
\label{fig:fmnist}
\end{figure*}
\section{Runtime and memory profiling of $\pi^2_{K}$ versus TEMP on software}\label{secA6}
To quantitatively evaluate the computational cost of the TEMP neuron relative to the proposed $\pi^2_{K}$ formulation, we conducted a profiling experiment in PyTorch on a CPU. We implemented a single fully connected layer using the TEMP rule and compared it against a $\pi^2_{K}$–based layer of identical dimensionality (input size $=1000$, output size $=100$). Both layers were initialized with identical, fixed random weights to ensure a controlled comparison, and randomly generated Gaussian input vectors ($N(0,1)$) of dimension $1000$ were used to measure peak memory consumption and execution time. For the TEMP layer, the parameter $\gamma$ was fixed at $10$, resulting in an average of 520 contributing inputs per output neuron, with the actual number varying between 500 and 550 across trials. Accordingly, we set $K=520$ in the $\pi^2_{K}$ neuron to provide a matched baseline. Averaged over ten runs, the $\pi^2_{K}$ neuron achieved a $6.5\times$ reduction in runtime and a $3\times$ reduction in memory usage compared to the TEMP neuron. These empirical improvements are consistent with the theoretical complexity difference: while TEMP requires a full sorting operation with $O(d \log d)$ cost, $\pi^2_{K}$ relies only on a partial sort (\texttt{topK}) with $O(d \log K)$ complexity.

\end{document}